%% file: main.tex
\documentclass{article}

     \usepackage[nonatbib, preprint]{neurips_2020}

\usepackage[utf8]{inputenc} 
\usepackage[T1]{fontenc}    
\usepackage{hyperref}       
\usepackage{url}            
\usepackage{booktabs}       
\usepackage{amsfonts}       
\usepackage{nicefrac}       
\usepackage{microtype}      
\usepackage{xcolor}
\usepackage{amsmath}
\usepackage{graphicx}
\usepackage{wrapfig}
\usepackage{multirow}
\usepackage{comment}
\usepackage{amsthm}
\usepackage{colortbl}
\usepackage{array}

\input{math_commands}

\def\showmainbody{1} 

\newcommand{\td}{\textnormal{TD}}

\newcommand{\cka}{\textnormal{CKA}}
\newcommand{\diag}{\textnormal{diag}}
\newcommand{\tr}{\textnormal{\textbf{tr}}}
\newcommand{\mspan}{\textnormal{span}}

\newtheorem{thm}{Theorem}
\newtheorem{lem}{Lemma}
\newtheorem{cor}{Corollary}

\newlength\savewidth
\newcommand\shline{\noalign{\global\savewidth\arrayrulewidth
                           \global\arrayrulewidth 1.2pt}%
                  \hline
                  \noalign{\global\arrayrulewidth\savewidth}}

\title{Transferred Discrepancy: Quantifying the Difference Between Representations}

\author{%
  Yunzhen Feng$^*$ \\
  Peking University \\
  \texttt{fengyz@pku.edu.cn} \\
   \And
  Runtian Zhai$^*$ \\
  Peking University \\
   \texttt{zhairuntian@pku.edu.cn} \\
   \AND
   Di He \\
   Microsoft Research\\
   \texttt{dihe@microsoft.com} \\
   \And
   Liwei Wang \\
  Peking University \\
   \texttt{wanglw@cis.pku.edu.cn} \\
   \And
   Bin Dong \\
  Peking University \\
   \texttt{dongbin@math.pku.edu.cn} \\
}

\begin{document}

\maketitle

\ifdefined\showmainbody{

\begin{abstract}
Understanding what information neural networks capture is an essential problem in deep learning, and studying whether different models capture similar features is an initial step to achieve this goal. Previous works sought to define metrics over the feature matrices to measure the difference between two models. However, different metrics sometimes lead to contradictory conclusions, and there has been no consensus on which metric is suitable to use in practice. In this work, we propose a novel metric that goes beyond previous approaches. Recall that one of the most practical scenarios of using the learned representations is to apply them to downstream tasks. We argue that we should design the metric based on a similar principle. For that, we introduce the transferred discrepancy (TD), a new metric that defines the difference between two representations based on their downstream-task performance.
Through an asymptotic analysis, we show how TD correlates with downstream tasks and the necessity to define metrics in such a task-dependent fashion.  In particular, we also show that under specific conditions, the TD metric is closely related to previous metrics. Our experiments show that TD can provide fine-grained information for varied downstream tasks, and for the models trained from different initializations, the learned features are not the same in terms of downstream-task predictions. We find that TD may 
also be used to evaluate the effectiveness of different training strategies. For example, we demonstrate that the models trained with proper data augmentations that improve the generalization capture more similar features in terms of TD, while those with data augmentations that hurt the generalization will not. This suggests a training strategy that leads to more robust representation also trains models that generalize better. 
\end{abstract}

\section{Introduction}

Deep neural networks have achieved great success in many real-world applications, such as image classification \cite{krizhevsky2012imagenet}, speech recognition \cite{hinton2012deep}, and natural language processing \cite{devlin2018bert}. It is generally agreed in the community that deep learning captures better representations than the previous hand-crafted feature engineering, which contributes to a significant performance improvement \cite{bhardwaj2018deep}. Therefore, it is worthwhile to investigate what features\footnote{Without any confusion, we use the terms \emph{feature} and \emph{representation} interchangeably.} a neural network learns in practice, which helps us understand the nature of deep learning. As an initial step towards this challenging task, many people study how different the features learned by different neural networks are.

Recent works \cite{2015arXiv151107543L, NIPS2017_7188, NIPS2018_8167, NIPS2018_7815, pmlr-v97-kornblith19a, Liang2020Knowledge} have proposed several metrics measuring the difference between a pair of features learned by two different models. Quite confusingly though, two metrics that both seem reasonable can even draw opposite conclusions on the same issue. For example, \cite{NIPS2018_8167} used the maximum match under linear transformations as the evaluation metric and found that the representations of two neural networks trained from different random initializations were utterly different. In sharp contrast, \cite{pmlr-v97-kornblith19a} measured with the central kernel alignment (CKA) and concluded that the models capture almost identical features. The contradiction caused by using different metrics is rooted in the disagreement towards the following question:

\begin{center}
\textit{What does it mean by two representations being different?}
\end{center}

Existing metrics measure the difference directly by feature values, which we believe is debatable. In representation learning, the quality of features learned by a neural network is hardly evaluated on their values, but rather on the performances they achieve when applied to downstream tasks \cite{devlin2018bert, kolesnikov2019revisiting, chen2020simple}. Given a trained feature extractor, we train an additional output head on the features of the data for each downstream task and consider one feature (or feature extractor) better than the other if it achieves higher downstream-task performance. Such a way of evaluation has been widely adopted and proven useful in computer vision \cite{chen2020simple} and natural language processing \cite{devlin2018bert}. 

Based on the above discussion, we argue that a reasonable representation difference metric should also be designed by incorporating the downstream tasks. To achieve this, we propose a new metric which takes the downstream task as an input, and refer to it as the transferred discrepancy (TD). Given two feature extractors and a set of downstream tasks, we train an output head on top of each feature extractor per task, and define the TD metric as the difference between the predictions over the downstream task data. The more different the two feature extractors are, the more likely they may lead to different predictions on the same downstream task data, and the higher TD value they will achieve. We analyze the theoretical properties of the TD metric under the linear probing setting, where the downstream tasks are limited to linear regression. Under this setting, we prove that the TD metric is invariant under reasonable transformations and analyze its asymptotic limits. Furthermore, we show that by properly selecting downstream tasks, the TD metric is closely related to existing metrics such as the maximum match \cite{NIPS2018_8167}, canonical correlation analysis (CCA)\cite{NIPS2017_7188}, and CKA\cite{pmlr-v97-kornblith19a}. 

Apart from theoretical analysis, we conduct extensive experiments and demonstrate that the TD metric can reveal faithful information for various downstream tasks in practice. Importantly, we observe that the features learned by models trained from different initializations are not quite the same according to their performances on downstream tasks. Furthermore, we study a quantity called \emph{TD robustness}, which is defined as the difference between the predictions of two models trained from different initializations on the same downstream task. We investigate how TD robustness varies upon changes in factors in deep learning such as data augmentation methods and training strategies. Remarkably, we find that TD robustness is closely connected with the quality of the representation. For instance, data augmentation methods that are proven to improve the quality of the representation \cite{shorten2019survey, chen2020simple} are also observed to increase TD robustness while transformations that harm the quality lower it. Such a relationship is also observed for other factors, providing a new perspective on how various factors in deep learning affect the representation the model learns.




\section{Related Work}

Many previous works try to understand whether two neural networks with drastically different parameters but similar high performances learn similar representations \cite{2015arXiv151107543L,NIPS2017_7188,NIPS2018_8167,NIPS2018_7815,pmlr-v97-kornblith19a, Liang2020Knowledge}. Using a neural network's hidden states as features, these works obtain a feature matrix over a set of samples for each model, and evaluate the correlation between the two feature matrices with some metrics taken from matrix theory. For example, \cite{NIPS2018_8167} measured the size of the intersection between two matrices' row spaces. \cite{NIPS2017_7188} applied CCA to large singular vectors of the two matrices, which is further improved by \cite{NIPS2018_7815}. \cite{pmlr-v97-kornblith19a} computed the norms of the generalized cross-correlation operator between two matrices. However, all of these works cast the quantification of the difference between two models as a matrix correlation problem, while the practical usage of the feature is largely ignored.

The proposed TD is motivated by representation learning. In representation learning, the quality of the learned representations is often evaluated based on their performance on downstream tasks. In computer vision, the representation of images can be pre-trained using SimCLR \cite{chen2020simple}, and such learned representations can help the model training on twelve downstream classification tasks. Another widely known milestone in natural language processing is the BERT model \cite{devlin2018bert}. Thus, we think it is more reasonable to define whether two representations are similar based on their performances on downstream tasks as well. 

\section{Transferred Discrepancy (TD)}\label{sec:theory}

Let $\gX$ be the input space, and $\mathbf x_1,\cdots,\mathbf x_n$ be $n$ i.i.d. samples from an underlying distribution $\pdata$ defined on $\gX$. Let $\Phi(\cdot)$ and $\Phi'(\cdot)$ be the two feature extractors, typically neural networks. Let $\mathbf z_i = \Phi(\mathbf x_i) \in \R^{p}$ and $\mathbf z_i' = \Phi'(\mathbf x_i)\in \R^{p'}$ be the feature of $\mathbf x_i$ extracted by $\Phi(\cdot)$ and $\Phi'(\cdot)$ respectively, $i=1,\cdots,n$. Note that
the dimensions of the features, $p$ and $p'$, are not necessarily equal, and we assume that $p \leq p'$ without loss of generality. Denote $Z=(\mathbf z_1,\cdots,\mathbf z_n) \in \R^{p \times n}$ and $Z'=(\mathbf z_1',\cdots,\mathbf z_n') \in \R^{p' \times n}$ as the feature matrices. Our goal is to quantify the difference between the two feature extractors $\Phi(\cdot)$ and $\Phi'(\cdot)$. 


As previously discussed, in representation learning, the feature extractor is designed and trained to improve the model's performance on downstream tasks. Thus, we argue that the difference between $\Phi(\cdot)$ and $\Phi'(\cdot)$ should be evaluated based on their corresponding performances on downstream tasks. In practice, a downstream task can be specified by a label vector $Y=(y_1,\cdots,y_n)$, where $y_i$ is the label of $\mathbf x_i$. $y_i$ can be a categorical value for classification tasks or a numeric value for regression tasks. Let $h_{W}(\mathbf z_i)$ and $h'_{W'}(\mathbf z'_i)$ be the output heads built upon the two feature extractors, $W$ and $W'$ are the learnable parameters. Given a loss function $\ell(\hat{y},y)$, the parameters $W$ and $W'$ are obtained by minimizing the empirical risks:

\begin{equation}\label{equ:argmin_weight}
\left \{
\begin{aligned}
\hat{W} & = \argmin_{W} \frac{1}{n} \sum_{i=1}^n \ell(h_{W}(\mathbf z_i), y_i), \\ 
\hat{W'} &= \argmin_{W'} \frac{1}{n} \sum_{i=1}^n \ell(h'_{W'}(\mathbf z'_i), y_i).
\end{aligned}
\right .
\end{equation}
After $\hat{W}$ and $\hat{W'}$ are obtained, the difference between $Z$ and $Z'$ is measured by the divergence between $h_{W}(Z)$ and $h'_{W'}(Z')$. Let $d(u,v)$ be a symmetric divergence function. The difference between $Z$ and $Z'$ on the task with label $Y$ is defined as\footnote{For ease of understanding, Eqn (\ref{equ:diff_metric}) defines the metric over the training data. Generally, one can also quantify this distance on unseen data, e.g., the test set in the downstream task, to take the generalization into account. In our experiments, we use TD metrics over the test data and find the value is similar to that over the training data.}:
\begin{equation}
\label{equ:diff_metric}
\td (Z, Z'; Y) = \frac{1}{n} \sum_{i=1}^n d(h_{\hat{W}}(\mathbf z_i) , h'_{\hat{W'}}(\mathbf z'_i)).
\end{equation}
As we quantify the difference between representations from a feature-transferring perspective, we name it \emph{transferred discrepancy}. The TD metric defined on one downstream task may be insufficient to measure the difference, so we further define the transferred discrepancy on a family of tasks. Suppose $\gS$ is a set of label vectors with each vector Y as a task. The TD measured on set $\gS$ is:
\begin{equation}
\label{equ:diff_metric_s}
    \td(Z, Z'; \gS) = \max_{Y \in \gS} \td(Z, Z'; Y).
\end{equation}
The TD metric can reveal the difference between $Z$ and $Z'$. Intuitively, when $Z$ and $Z'$ are very similar, $\hat W$ and $\hat W'$ will be similar and $h_{\hat W}(\mathbf z)$ and $h_{\hat W'}(\mathbf z)$ will not differ much from each other on every data in the downstream task. Thus, $Z$ and $Z'$ will have a small TD value. On the contrary, when $Z$ and $Z'$ are different features, the prediction $h_{\hat W}(\mathbf z)$ and $h_{\hat W'}(\mathbf z)$ are more likely different and the value of TD will be large. Although the TD metric is designed from a practitioner's perspective and is subject to the downstream tasks, we will show that it has nice theoretical properties and close relation with previous metrics in the next section.

\section{Theoretical Properties of the TD metric}\label{sec:property}
In this section, we theoretically analyze the TD metric under the \emph{linear probing} setting. We show that TD is invariant to orthogonal transformation and isotropic scaling (4.1), and further investigate its asymptotic behavior as $n$ approaches infinity (4.2). Finally, we demonstrate that under certain conditions, the TD metric and three previous metrics, maximum match, CCA, and CKA, depend on the same statistics (4.3).

\subsection{Transformation Invariance}

The linear probing setting is widely studied in literature \cite{alain2016understanding,oord2018representation,hjelm2018learning} and applied in practice \cite{chen2020simple,NIPS2019_9081}. In this setting, a linear model is trained on top of a feature extractor, and its performance is used as a proxy for the quality of the features. Define $h_W(\mathbf z) = W\mathbf z + b$ and $h'_{W'}(\mathbf z')=W' \mathbf z' + b'$, where $W\in \R^{p},W'\in \R^{p'}$ and $b,b'\in \R$. Let $\ell(\hat{y}, y) = (\hat{y}-y)^2$ be the square loss and $d(u, v) = (u-v)^2$ be the squared distance. Following \cite{pmlr-v97-kornblith19a}, we assume that both $Z$ and $Z^{\top}$ have been preprocessed to center the rows and $Y$ is centered\footnote{Note that this assumption only simplifies the proof. Without such a preprocessing, we can get similar results with more calculations.} . Besides, we assume that we have enough data such that $n > \max\{p, p'\}$, and both empirical covariance matrices $ZZ^{\top}$ and $Z'Z'^{\top}$ are invertible\footnote{In practice we have $n > \max\{p, p'\}$ in most if not all cases. Even if $ZZ^{\top}$ is not invertible, adding a tiny noise to $Z$ makes $ZZ^{\top}$ invertible.} . Under these assumptions, the optimization problem (\ref{equ:argmin_weight}) can be rewritten as
\begin{equation}
\label{equ:argmin_weight_rewrite}
\left \{
\begin{aligned}
\hat{W} , b& = \argmin_{W, b} \frac{1}{n} \sum_{i=1}^n (W\mathbf z_i+b- y_i)^2, \\ 
\hat{W'} , b'&= \argmin_{W', b'} \frac{1}{n} \sum_{i=1}^n (W'\mathbf z'_i+b'-y_i)^2.
\end{aligned}
\right .
\end{equation}
As $Z,Z'$ are all row-centered, problem (\ref{equ:argmin_weight_rewrite}) has a simple closed-form solution: $\hat{b} = \hat{b'} = 0$ and $\hat{W} = YZ^{\top}(ZZ^{\top})^{-1}, \quad \hat{W'} = YZ'^{\top}(Z'{Z'}^{\top})^{-1}$. Plugging this solution into Eqn (\ref{equ:diff_metric}) yields
\begin{equation}
\label{equ:td-linear}
\td(Z, Z'; Y) = \frac{1}{n} \left\| Y[Z^{\top}(ZZ^{\top})^{-1}Z - Z'^{\top}(Z'Z'^{\top})^{-1}Z'] \right\|_2^2.
\end{equation}
A reasonable metric should have some basic properties.  \cite{pmlr-v97-kornblith19a} proposed that a similarity metric of features should have two invariance properties, the invariance to isotropic scaling, and the invariance to orthogonal transformation. It is straightforward to see that (1). For any $\beta, \beta' \in \R^+$, $\td(Z,Z';Y)=\td(\beta Z,\beta' Z';Y)$. (2). For any unitary matrices $Q \in \R^{p \times p}$ and $Q' \in \R^{p' \times p'}$, $\td(Z,Z';Y)=\td(QZ,Q'Z';Y)$. Thus, Eqn (\ref{equ:td-linear}) derives that TD is invariant to isotropic scalings and orthogonal transformations. In fact, such transformation invariance holds not only for the square loss function but also for many other choices of $\ell$ as long as $\ell$ is strongly convex. We leave this discussion in the supplementary material.

\subsection{Convergence Analysis}
Since $\mathbf x_1,\cdots,\mathbf x_n$ are independently sampled from $\pdata$, it is natural to require that as $n$ goes to infinity, $\td(Z, Z'; \gS)$ converges to a value which represents the difference of two feature extractors over the data distribution on downstream task set $\gS$. We denote this value by $\td(\Phi(\pdata), \Phi'(\pdata); \gS)$. Denote the joint feature distribution for $Z$ and $Z'$ as $(P, P'):=(\Phi(\pdata), \Phi'(\pdata))$ where $P$ and $P'$ are corresponding marginal distributions. Since the rows of $Z$ and $Z'$ are centered, we have $\E_{\mathbf z \sim P}[\mathbf z] = 0$ and $\E_{\mathbf z' \sim P'}[\mathbf z'] = 0$. Let the covariance matrix of the joint distribution $(P, P')$ be $\begin{bmatrix} A&B\\B^{\top}&C \end{bmatrix}$. We also assume that the covariance matrices $A$ and $C$ are invertible. 
Since we seek to define TD on downstream tasks, the following \textit{representative task set} contains all possible tasks associated with $Z$ and $Z'$. All proofs can be found in the supplementary material.

\textbf{Representative Task Set  } 
Define $\gS^* = \{Y = \bm \alpha A^{-\frac{1}{2}}Z + \bm \alpha' C^{-\frac{1}{2}} Z' :\bm \alpha \in \R^{p},\bm \alpha \in \R^{p'}, \left\| \bm \alpha \right\|_2 \leq 1, \left\| \bm \alpha' \right\|_2 \leq 1\}$. This set contains all tasks linearly realizable by $Z$ and $Z'$ thus covers a wide range of tasks. $A^{-\frac{1}{2}}$ and $C^{-\frac{1}{2}}$ are used to normalize $Z$ and $Z'$ so that they have the same level of contribution to $\gS^*$. For any downstream task $Y$ in $\gS^*$, the following theorem rigorously states how the TD metric depends on $Y$ and the representations $Z$, $Z'$.
\begin{thm}
\label{thm:td-1task}
Suppose $A,B$, and $C$ are defined as above. Denote $D=A^{-\frac{1}{2}}BC^{-\frac{1}{2}}$. For downstream task $ Y=\bm \alpha A^{-\frac{1}{2}}Z+\bm \alpha' C^{-\frac{1}{2}} Z'$, we have
\begin{equation}
\label{equ:td-1task}
\td(Z,Z'; Y) \xrightarrow{a.s.} \ \bm \alpha(I_p - DD^{\top})\bm \alpha^{\top} +\bm \alpha'(I_{p'} - D^{\top}D)\bm \alpha'^{\top}+ 2\bm \alpha(D-DD^{\top}D)\bm \alpha'^{\top},
\end{equation}
\end{thm}
Theorem \ref{thm:td-1task} indicates that in this setting, the transferred discrepancy primarily depends on the task (i.e., $\alpha$ and $\alpha'$) and the matrix $D$ which leverages the covariance matrix $B$ between $Z$ and $Z'$. Here we show two special cases where Eqn \ref{equ:td-1task} has a simpler form for better understandings:

\begin{itemize}
    \item TD for linearly correlated features: If $p=p'$ and there exists a unitary matrix $Q \in \R^{p \times p}$ such that $(z,z')$ drawn from $(P,P')$ satisfies $z'=Qz$, i.e., $Z$ and $Z'$ are linearly correlated. We have $D=Q^{\top}$, so $\td(Z, Z'; \bm \alpha A^{-\frac{1}{2}}Z+\bm \alpha' C^{-\frac{1}{2}} Z') \xrightarrow{a.s.} 0$.
    \item TD for independent features: If $P$ and $P'$ are independent, then $B=0$ and $D=0$. Consequently, $\td(Z, Z'; \bm \alpha A^{-\frac{1}{2}}Z+\bm \alpha' C^{-\frac{1}{2}} Z') \xrightarrow{a.s.} 2$.
\end{itemize}
Next, we provide the convergence analysis of the TD metric on a set of tasks. We first study the asymptotic limit for TD on the \textit{representative task set}, and then extend the analysis to the \textit{restricted task sets}. Both theoretical results show that TD highly relates to the singular value distribution of matrix $D$. 

\begin{thm}
\label{thm:convergence}
Under the notations in Theorem \ref{thm:td-1task}, denote $\sigma_1 \geq \cdots \geq \sigma_p \geq 0$ be the singular values of $D$. If $p = p'$, or $p < p'$ and $(1-\sigma_p)(1+\sigma_p)^2 \geq 1$, we have a closed-form limit of TD on the representative task set $\gS^*$:
\begin{equation}\label{equ:repre_conv}
    \td(Z, Z'; \gS^*) \xrightarrow{a.s.} \max_{j = 1,\cdots,p} 2(1-\sigma_j)(1+\sigma_j)^2.
\end{equation}
\end{thm}
The result shows that the TD metric on the representative task set is closely related to the singular values of $D$. For example, if $\sigma_p$ is small, there exists a task $Y\in \gS^*$ that is close to the row space of one feature matrix but nearly orthogonal to the row space of the other. Two representations will have drastically different performance on Y, and $\td(Z, Z'; \gS^*)$ is large. 
If $\sigma_p$ is close to 1, so do all the singular values, and any task $Y\in \gS^*$ will be close to both the row spaces of $Z$ and $Z'$. Thus, the performance is similar, and $\td(Z, Z'; \gS^*)$ is small. When there are both large and small singular values, the difference between the performance varies a lot on diverse tasks, and evaluating the difference on specific tasks may be helpful. The following corollary gives a convergence guarantee for a more practical case for subsets of the representation task set:

\textbf{Restricted Task Set  } 
We study smaller task sets called restricted task sets that allow us to jointly consider $\sigma_1,\cdots,\sigma_p$ instead of $\sigma_p$ alone. Particularly, we construct a cascade of sets $\gS_1 \subset \cdots \subset \gS_p \subset \gS$, and the two representations are considered more similar if they have similar performance on a larger task set. Define $\gS_r = \{Y = \bm \alpha A^{-\frac{1}{2}}Z + \bm \alpha' C^{-\frac{1}{2}} Z' : \left\| \bm \alpha \right\|_2 \leq 1, \left\| \bm \alpha' \right\|_2 \leq 1, \bm \alpha \mathbf u_j = 0 \text{ and } \bm \alpha' \mathbf v_j= 0 \text{ if } j > r\}$, $r=1,\cdots,p$, where $\mathbf u_j$ and $\mathbf v_j$ are the $j$-th singular vectors of $U, V$, in $D=U\Sigma V^{\top}$ respectively. When $p=p'$, we have $\gS_p = \gS^*$. The following corollary gives the asymptotic limits of the TD metric on the restricted task sets:

\begin{cor}
\label{thm:td-restricted}
Under the notations in Corollary \ref{thm:convergence} and for the restricted task set $\gS_r$ defined as above, we have
\begin{equation}
\label{equ:td-restricted}
    \td(Z, Z'; \gS_r) \xrightarrow{a.s.} \max_{j=1,\cdots,r}2(1-\sigma_j)(1+\sigma_j)^2.
\end{equation}
\end{cor}

We conduct experiments to empirically analyze the singular values of $D$ for some $Z$ and $Z'$ trained on the Cifar-10 dataset \cite{Krizhevsky2009Learning}. All the results can be found in the supplementary material. 

\subsection{Connection With Previous Metrics}
\label{sec:cca-d}

In this section, we show that our TD metric has a close connection with previous metrics by showing the matrix $D$ defined in section 4.2 is also the key to derive previous metrics. Previous works lead to contradictive conclusions by using $D$ in different ways. We only discuss the relationship with CCA in the main body and leave the discussions on maximum match and CKA in the supplementary material.

\textbf{Canonical Correlation Analysis (CCA)  } CCA measures the relationship between $Z$ and $Z'$ by finding two bases of their row spaces such that when projected onto these bases, the correlation between the two matrices is maximized. Formally, for $1\le j\le p$, define the maximum correlation coefficient $\rho_i$ by the following optimization problem:
\begin{equation}
\begin{aligned}
\rho_{j} = \max_{\mathbf{w}_j, \mathbf{w}'_j}\quad & \frac{\text{cov}(\mathbf{w}_j^{\top}Z, \mathbf{w}'^{\top}_jZ')}{\sqrt{\text{var}(\mathbf{w}_j^{\top}Z)\text{var}(\mathbf{w}'^{\top}_jZ')}},\\
\text{subject to} \quad & \text{cov}(\mathbf{w}_k^{\top}Z, \mathbf{w}_j^{\top}Z)=0, \text{ cov}(\mathbf{w}'^{\top}_kZ', \mathbf{w}'^{\top}_jZ')=0,  \forall k<j.
\end{aligned}
\end{equation}
Then the summary statistics of CCA is defined as $R_{CCA}^2 := \frac{\sum_{j=1}^p \rho_j^2}{p}$.
Equivalently, let $Q=Z^{\top}(ZZ^{\top})^{-\frac{1}{2}}$ and $Q'=Z'^{\top}(Z'Z'^{\top})^{-\frac{1}{2}}$, then $R_{CCA}^2 = \frac{\left\|Q'^{\top}Q \right\|_F^2}{p}$. Since data are independently sampled from the distribution, we have $\E_{Z,Z'}[Q'^{\top}Q]=D^{\top}$. By $\left\|D\right\|_F^2=\sum_{j=1}^p \sigma_j^2$, we have:
\begin{equation} \label{equ:cca_sing}
\E_{Z,Z'}[R_{CCA}^2]=\frac{\sum_{j=1}^{p}\sigma_j^2}{p}.
\end{equation}
Moreover, take $\hat A = \frac{ZZ^{\top}}{n}$ and $\hat C = \frac{Z'Z'^{\top}}{n}$ as the empirically estimators of $A$ and $C$. CCA is directly related to TD on an empirical representative task set $\hat \gS^*$ with $Y=\bm \alpha \hat A^{-\frac{1}{2}}Z+\bm \alpha' \hat C^{-\frac{1}{2}}Z'$ as below:

\begin{thm}
\label{thm:mean_convergence}
Let $\alpha$ and $\alpha'$ be uniformly distributed on the unit ball and $\hat A, \hat C$ defined above. If $p=p'$,
\begin{equation}
\label{equ:repre_conv_cca}
\E_{\alpha, \alpha'}[\td(Z,Z'; Y=\bm \alpha \hat A^{-\frac{1}{2}}Z+\bm \alpha' \hat C^{-\frac{1}{2}}Z')]=\frac{2p}{p+2}(1-R_{CCA}^2).
\end{equation}
\end{thm}
This result reveals the connection between CCA and our proposed metric. CCA is equivalent to averaging TD on all linearly realizable tasks in the linear probing setting, which makes CCA downstream-task agnostic. However, when evaluated on a subset of tasks, or several tasks of interest, CCA cannot provide fine-grained information, while the TD metric is always faithful to the downstream tasks. In the next section, we conduct extensive experiments to study the performance of the TD metric in practical applications.

\section{Experiments}
\label{sec:experiments}
In this section, we empirically compare one TD induced metric, $\td_{cls}$, with CCA and CKA for classification tasks. For convenience, we define the three metrics as below:
\begin{itemize}
    \item $D_{CCA}(Z,Z') = 1 - R_{CCA}^2$
    \item $D_{CKA}(Z,Z') = 1 - S_{CKA} (Z^{\top}Z, Z'^{\top}Z')=1-\frac{\|ZZ'^{\top}\|_F^2}{\|ZZ^{\top}\|_F\|Z'Z'^{\top}\|_F}$ 
    \item $\td_{cls}(Z, Z';Y)=\frac{1}{n}\sum_{i=1}^n \mathbf{1}_{\argmax h(\mathbf z_i) = \argmax h'(\mathbf z'_i)}$
\end{itemize}
where $\argmax h(\mathbf z)$ denotes the output prediction for $\mathbf z$. It is easy to see that $\td_{cls}$ measures the fraction of different predictions between two representations on the downstream task $Y$. We also try some other distance metrics and obtain similar conclusions. The results are in the supplementary material. We will firstly show that the TD metric is valid with a sanity check, and then 
investigate whether models that are trained using different initializations learn similar features. Lastly, we will define TD robustness and study the TD robustness of different training strategies. We will also establish a connection between TD robustness and the quality of the representation.

\subsection{A Sanity Check}
\label{exp:sanity}

In this section, we demonstrate the validity of the TD metric by a simple sanity check: given three feature extractors $\Phi_1, \Phi_2$, $\Phi_3$, such that $\Phi_1$ and $\Phi_2$ capture similar features than $\Phi_1$ and $\Phi_3$ by some prior knowledge, we check whether the value of the TD metric is consistent with the prior. 

To build this sanity check, we train $\Phi_1$ and $\Phi_2$ using similar datasets, but train $\Phi_3$ using a completely different dataset. Specifically, we design Cifar-2 and Cifar-5 by grouping Cifar-10's labels into 2 and 5 groups respectively. Therefore, Cifar-2, Cifar-5, and Cifar-10 datasets share the same input data but use slightly different labels. 
These three datasets serve as candidates for training $\Phi_1$ and $\Phi_2$. $\Phi_3$ is trained using the SVHN dataset. All feature extractors use ResNet32 architecture, and the classification head is removed after training. For simplification, we use the Cifar-10 task as the downstream task. We apply $\Phi_1$, $\Phi_2$, $\Phi_3$ to the training data, and train a classification head for Cifar-10 using logistic regression. After training, we apply the three classifiers to the test set and calculate $\td_{cls}$. For a fair comparison, both $D_{CCA}$ and $D_{CKA}$ are also calculated on the test samples.  Each experiment is conducted for ten times. All details can be found in the supplementary material.
 
We compare the difference between pairs of models using different metrics and list all results in Table \ref{tab:cifarvssvhn}. It can be seen that both $D_{CCA}$ and $D_{CKA}$ have smaller values in the first three rows compared with the other rows, which indicates that the features learned from Cifar-2/5/10 are more similar, but are very different from the features learned from SVHN. $\td_{cls}$ also has a similar trend: the values of $\td_{cls}$ in the first three rows are smaller than 0.5 while the values in the other rows are all larger than 0.8. Thus, all these three metrics are reasonable and consistent with our prior knowledge.

\begin{table}[htb]
\centering
\vspace{.05in}
\caption{The sanity check result.}
    \begin{tabular}{|l|l|l|l|l|l|l|}
    \hline
        Model 1 & Model 2 & Downstream & $D_{CCA}$ & $D_{CKA}$ & $\td_{cls}$ \\ \hline
        \rowcolor[RGB]{234,255,255}Cifar-10 & Cifar-5 &  & \cellcolor[RGB]{251, 158, 147}0.7642 & \cellcolor[RGB]{253, 217, 212}0.3024  & \cellcolor[RGB]{254, 228, 225}0.2158  \\ \cline{1-2}\cline{4-6}
        \rowcolor[RGB]{234,255,255} Cifar-10 & Cifar-2 & Cifar-10 & \cellcolor[RGB]{251, 149, 138}0.8323 & \cellcolor[RGB]{252, 187, 180}0.5330 & \cellcolor[RGB]{253, 192, 185}0.4958  \\ \cline{1-2}\cline{4-6}
        \rowcolor[RGB]{234,255,255} Cifar-5 & Cifar-2 & & \cellcolor[RGB]{251, 153, 142}0.8030 & \cellcolor[RGB]{253, 197, 191}0.4530 & \cellcolor[RGB]{253, 192, 185}0.4944 \\ \cline{1-2}\cline{4-6}
        \rowcolor[RGB]{235, 235, 235}Cifar-10 & SVHN &  & \cellcolor[RGB]{250, 138, 125}0.9218 & \cellcolor[RGB]{250, 132, 118}0.9713 & \cellcolor[RGB]{251, 153, 142}0.8001 \\ \cline{1-2}\cline{4-6}
        \rowcolor[RGB]{235, 235, 235}Cifar-5 & SVHN & Cifar-10 & \cellcolor[RGB]{250, 131, 117}0.9793 & \cellcolor[RGB]{250, 132, 118}0.9710 & \cellcolor[RGB]{251, 153, 142}0.8049  \\ \cline{1-2}\cline{4-6}
        \rowcolor[RGB]{235, 235, 235}Cifar-2 & SVHN & & \cellcolor[RGB]{250, 139, 126}0.9163 & \cellcolor[RGB]{250, 131, 117}0.9782 & \cellcolor[RGB]{251, 151, 140}0.8163  \\ \hline
    \end{tabular}
\vspace{-.15in}
    \label{tab:cifarvssvhn}
\end{table}

\subsection{Does Initialization Affect Learned Features?}
\label{exp:moreinfo}

In this section, we address the problem widely studied by previous works \cite{NIPS2018_8167, pmlr-v97-kornblith19a, NIPS2018_7815}: whether models trained from different random initializations learn different features or not. We train two ResNet32 networks on the training set of Cifar-5/2 using the same setting (dataset, algorithm, hyperparameters, etc.) but with different random initializations. We use three downstream tasks, Cifar-10/5/2. $D_{CCA}$ $D_{CKA}$ are directly computed over the test set. For TD, we first train the output heads of the models on the training data and then compute $\td_{cls}$  on the test data. The results are reported in Table \ref{tab:diffseed}. For each row, we repeat the experiment ten times and report the average values.

\begin{table}[htb]
    \centering
    \caption{Similarity of the models learned with different initializations.}
    \begin{tabular}{|l|l|l|l|l|l|l|}
    \hline
        Model 1& Model 2  & Downstream & $D_{CCA}$ & $D_{CKA}$ & $\td_{cls}$  \\ \hline
        \multirow{3}{*}{Cifar-5} & \multirow{3}{*}{Cifar-5} & Cifar-10 & \multirow{3}{*}{0.6961} & \multirow{3}{*}{0.0835} & 0.2139  \\ \cline{3-3} \cline{6-6}
         & & Cifar-5 & & & 0.0442  \\ \cline{3-3} \cline{6-6}
         & & Cifar-2 & & & 0.0109 \\ \hline
        \multirow{3}{*}{Cifar-2} & \multirow{3}{*}{Cifar-2} & Cifar-10 & \multirow{3}{*}{0.6931} &  \multirow{3}{*}{0.0402} & 0.3745 \\ \cline{3-3}\cline{6-6}
         & & Cifar-5 & & & 0.2631\\ \cline{3-3}\cline{6-6}
         & & Cifar-2 & & & 0.0164  \\ \hline
    \end{tabular}
    \label{tab:diffseed}
\end{table}

It can be seen that $\td_{cls}$ provides more information than CCA and CKA. Since CCA and CKA are downstream-task-agnostic, they can only present one scalar value for each pair of models. However, $\td_{cls}$ outputs different values for different downstream tasks. For example, the two models trained on Cifar-5 have very consistent predictions on Cifar-2 and Cifar-5, but their predictions are much more different on Cifar-10: they disagree on 21.39\% of the Cifar-10 test samples. It can be also observed that the two models trained on Cifar-2 behave similarly on the Cifar-2 downstream task but quite differently on Cifar-5 and Cifar-10. 

We find that the two feature extractors trained from different initializations do not learn the same features. First of all, we show that our experimental setting is reasonable since the downstream tasks have a close connection with the upstream tasks. To verify this, we find that the model trained from the Cifar-5 task can reach 80\% accuracy on the Cifar-10 downstream task. This result indicates that for those models, the features can be transferred from one task to the other. However, in this reasonable setting, the disagreements between representations are high, i.e., 21.39\% for two Cifar-5 models and 37.45\% for two Cifar-2 models. Such difference indicates that the features of these models are not the same when evaluated on Cifar-10. Our finding is consistent with the results in \cite{NIPS2018_7815} that models trained with different initializations can capture different features, but differs from the results in \cite{pmlr-v97-kornblith19a}.

\subsection{TD Robustness and Its Application to Training Strategy Evaluation}
\label{exp:sim-better}

Deep learning practitioners design different training strategies in the hope of enhancing the quality of features captured by the model. In this section, we build a connection between the quality of the learned representations and the \emph{TD robustness} of the implemented training strategy. A training strategy is said to be TD-robust if it leads to models with consistent predictions on the same downstream task when trained from different initializations. We study the effect of different factors on TD robustness. Specifically, we focus on three factors: data augmentation, learning rate schedules and adversarial training. Experimental details are left in the supplementary material.

\textbf{Data Augmentation} \quad
We study three augmentation methods: random flipping, random cropping, and adding Gaussian noise. Random flipping and cropping are widely used in practice and believed helpful to generalization\cite{shorten2019survey}. Gaussian additive noise is used for learning smooth classifiers \cite{cohen2019certified}. 

\begin{table}[htb]
    \centering
    \caption{The effect of difference factors in training. For each of the three factors, the bold character indicates the configuration with the best TD robustness.}
    \begin{tabular}{lllll}
    \shline
    \specialrule{0em}{1.2pt}{1.2pt}
      Factors  & Configuration & Upstream & Downstream & $\td_{cls}$ \\
      \specialrule{0em}{1.2pt}{1.2pt}
        \hline
        \specialrule{0em}{1.2pt}{1.2pt}
        \multirow{5}{*}{Data Augmentation} & Without augmentation & \multirow{5}{*}{Cifar-5} & \multirow{5}{*}{Cifar-10} & 0.2812\\
        & Random flipping & & & 0.2739 \\
        & Random flipping + croping & & & \textbf{0.2150} \\
        & Random flipping + croping  & & & \multirow{2}{*}{0.2542}  \\
        & + Gaussian additive noise & & & \\
        \cline{2-5}
        \specialrule{0em}{1.2pt}{1.2pt}
        \multirow{3}{*}{LR Schedule} & Small LR + without decay & \multirow{3}{*}{Cifar-5} & \multirow{3}{*}{Cifar-10} & 0.2521 \\
        & Large LR + without decay & & & 0.2770 \\
        & Large LR + with decay & & & \textbf{0.2313}\\ \cline{2-5}
        \specialrule{0em}{1.2pt}{1.2pt}
        \multirow{2}{*}{Std / Adv training} & Standard Training & \multirow{2}{*}{Cifar-5} & \multirow{2}{*}{Cifar-10} & 0.2150\\
        & Adversarial Training & & & \textbf{0.1823}\\
        \shline
    \end{tabular}
    \label{tab:combine}
\end{table}

We report the results in Table \ref{tab:combine}. It can be seen that using cropping and flipping leads to more similar representations, and cropping has a significant impact on $\td_{cls}$. On the other hand, adding Gaussian noise has the opposite effect and makes representations less similar. This result aligns well with previous works that show random flipping and cropping can improve the quality of the representation, but Gaussian noise hurts the quality \cite{chen2020simple}.

\textbf{Learning Rate Schedule} \quad
Practically, a learning rate decay scheduler is usually used for the sake of finding the local minima. We here show that models using a learning-rate decay schedule also learn more similar features across different initializations. The result in Table \ref{tab:combine} shows that the "Large LR + with decay" schedule leads to more similar representations than the other two schedules. This observation coincides with the fact that learning rate decay helps improve the quality of the representation in practice, and validates the theoretical finding in \cite{NIPS2019_9341}.


\textbf{Adversarial Training} \quad
Recently, \cite{tsipras2019robustness, ilyas2019adversarial, engstrom2019learning} showed that adversarial training helps neural networks learn features that align better with human perceptions and \cite{NIPS2019_8409} used adversarially trained network to boost performance in synthesis tasks. We find that adversarial training is more TD-robust than standard training, as shown in Table \ref{tab:combine}. Our result suggests that adversarial training may capture features with better qualities even though it lowers the accuracy.

\textbf{TD Robustness Versus the Quality of the Representation} \quad
Our experimental results demonstrate a strong connection between TD robustness and the quality of the representation: a training strategy leading to better qualities produces better TD robustness. This connection makes TD a useful tool for investigating the effect of different factors on learned representations and designing new training strategies. We further discuss this connection in the supplementary material. 

\section{Conclusion and Future Work}

In this work, we propose the Transferred Discrepancy, a metric that quantifies the difference between two representations using the difference between their performance on the same set of downstream tasks. Our rigorous theoretical analysis founds a solid basis for TD and reveals the connection between TD and previous metrics. We also conduct extensive experiments to study how different training factors affect the difference between the two representations trained from different random initializations. We believe that using downstream-task performances to study how different the features learned by different models are is promising, and in the future we will study the effect of more factors such as normalization and dropout. At the same time, we would also like to extend the current setting to study the theory of deep transfer learning.


\label{sec:conclusion}

\section*{Broader Impact}

The Transferred Discrepancy (TD) proposed in this work not only serves as a measurement of difference between the representations learned by two neural networks, but also provides a new angle of understanding deep learning and assessing the effect of different factors on deep learning. We believe that TD can help the community in the following ways: (i) it helps people rethink how to define the difference between representations, and more importantly it implies that this difference should depend on specific downstream tasks; (ii) the connection between higher TD robustness and better quality of the representation suggests that we can make neural networks learn better features by enforcing higher TD robustness. Therefore, TD is a useful tool for building advanced models and boosting neural networks' performance in the real-world. Further theoretical analysis of the connection between TD robustness and quality of features may also provide new insight into learning representation.

\bibliographystyle{plain}  
\bibliography{references}  

}\fi

\newpage
\appendix

\section{Asymptotic Limits of Transferred Discrepancy}

\subsection{Proof of Theorem \ref{thm:td-1task}}

Suppose the covariance matrix of the joint distribution is $(\Phi(p_{data}), \Phi'(p_{data}))$. By the law of large numbers, as $n \rightarrow \infty$, $\frac{ZZ^{\top}}{n} \rightarrow A$, $\frac{ZZ'^{\top}}{n} \rightarrow B$ and $\frac{Z'Z'^{\top}}{n} \rightarrow C$ almost surely. Denote $D=A^{-\frac{1}{2}}BC^{-\frac{1}{2}}$. For downstream task $ Y=\bm \alpha A^{-\frac{1}{2}}Z+\bm \alpha' C^{-\frac{1}{2}} Z'$, we have

\begin{equation}
\begin{aligned}
\td(Z,Z'; Y) \text{   }= & \frac{1}{n} \left\| \bm \alpha A^{-\frac{1}{2}} [ZZ^{\top}(ZZ^{\top})^{-1}Z - ZZ'^{\top}(Z'Z'^{\top})^{-1}Z'] \right.\\
&\left. +\bm \alpha' A^{-\frac{1}{2}} [Z'Z^{\top}(ZZ^{\top})^{-1}Z - Z'Z'^{\top}(Z'Z'^{\top})^{-1}Z'] \right\|_2^2\\
\xrightarrow{a.s.}& \frac{1}{n} \left\| \bm \alpha A^{-\frac{1}{2}}(Z - BC^{-1}Z') + \bm \alpha '  C^{-\frac{1}{2}}(B^{\top}A^{-1}Z - Z') \right\|_2^2 \\
=& \frac{1}{n} \left[\left\|\bm \alpha A^{-\frac{1}{2}}(Z - BC^{-1}Z') \right\|_2^2 + \left\|\bm \alpha '  C^{-\frac{1}{2}}(B^{\top}A^{-1}Z - Z') \right\|_2^2 \right.\\ 
    &+ \left. 2 \bm \alpha A^{-\frac{1}{2}}(Z-BC^{-1}Z')(B^{\top}A^{-1}Z-Z')^{\top}C^{-\frac{1}{2}}\bm \alpha'^{\top} \right] \\ 
\xrightarrow{a.s.} & \bm \alpha(I_p - DD^{\top})\bm \alpha^{\top} +\bm \alpha'(I_{p'} - D^{\top}D)\bm \alpha'^{\top}+ 2\bm \alpha(D-DD^{\top}D)\bm \alpha'^{\top},
\end{aligned}
\label{equ:suppthm1}
\end{equation}

Thus, we have the results in Theorem \ref{thm:td-1task}. It shows that the difference between $Z$'s and $Z'$'s performance primarily depends on the matrix $D$ and the task $Y$ ($\bm \alpha, \bm \alpha'$). \qed

\subsection{Proof of Theorem \ref{thm:convergence}}

When evaluated on the \emph{representative task set}, $\gS^* = \{Y = \bm \alpha A^{-\frac{1}{2}}Z + \bm \alpha' C^{-\frac{1}{2}} Z' :\bm \alpha \in \R^{p},\bm \alpha \in \R^{p'}, \left\| \bm \alpha \right\|_2 \leq 1, \left\| \bm \alpha' \right\|_2 \leq 1\}$, the transferred discrepancy is

\begin{equation}
\label{equ:limsup}
    \begin{aligned}
    \td(Z, Z'; \gS^*) \text{  } = &  \sup_{\|\bm \alpha\|_2 \le 1, \|\bm \alpha'\|_2\le 1} \td(Z,Z'; Y\bm \alpha A^{-\frac{1}{2}}Z + \bm \alpha' C^{-\frac{1}{2}} Z')\\
    \xrightarrow{a.s.} &\lim_{n\rightarrow\infty}\sup_{\|\bm \alpha\|_2 \le 1, \|\bm \alpha'\|_2\le 1} \td(Z,Z'; Y\bm \alpha A^{-\frac{1}{2}}Z + \bm \alpha' C^{-\frac{1}{2}} Z').
\end{aligned}
\end{equation}

The TD metric is a polynomial function of $\bm \alpha$ and $\bm \alpha'$ and the limit only affect the coefficients, and $\mathcal{A}=\{(\bm \alpha, \bm \alpha')| \|\bm \alpha\|_2 \le 1, \|\bm \alpha'\|_2\le 1\}$ is a compact set in $\R^{p+p'}$. Thus, the $\lim$ and the $\sup$ in Eqn (\ref{equ:limsup}) is interchangeable. According to Theorem \ref{thm:td-1task}, there is

\begin{equation}
    \begin{aligned}
    \td(Z, Z'; \gS^*) \text{  } \xrightarrow{a.s.} &\sup_{(\bm \alpha, \bm \alpha)\in \mathcal{A}} \lim_{n\rightarrow\infty} \td(Z,Z'; Y\bm \alpha A^{-\frac{1}{2}}Z + \bm \alpha' C^{-\frac{1}{2}} Z')\\
    =& \sup_{(\bm \alpha, \bm \alpha)\in \mathcal{A}} [\bm \alpha(I_p - DD^{\top})\bm \alpha^{\top} + \bm \alpha'(I_{p'} - D^{\top}D)\bm \alpha'^{\top} + 2\bm \alpha(D-DD^{\top}D)\bm \alpha'^{\top}].
\end{aligned}
\end{equation}

Let the singular values of $D$ be $\sigma_1\ge\cdots\ge \sigma_p$, and the SVD of $D$ be $D=U\Sigma V^{\top}$. Since the covariance matrix $\begin{bmatrix}A&B\\B^{\top}&C\end{bmatrix}\succeq0$, its schur complement $A-BC^{-1}B^{\top}\succeq0$. Thus,

\begin{equation}
A-BC^{-1}B^{\top}=A^{\frac{1}{2}}\left(I_p-A^{-\frac{1}{2}}BC^{-1}B^{\top}A^{-\frac{1}{2}}\right)A^{\frac{1}{2}}= A^{\frac{1}{2}}\left(I_p-DD^{\top}\right)A^{\frac{1}{2}} \succeq0,
\end{equation}
which implies that $1\ge \sigma_1\ge\cdots\ge \sigma_p\ge0$. Denote $\alpha U = \bm \beta = (\beta_1,\cdots,\beta_p)$ and $\alpha' V = \bm \beta' = (\beta'_1,\cdots,\beta'_{p'})$. Then $\left\|\bm \beta \right\|_2 \leq 1$, $\left\| \bm \beta' \right\|_2 \leq 1$, and

\begin{equation}
\label{equ:conv-mid}
\td(Z, Z'; \gS^*) \xrightarrow{a.s.} \sup_{\bm \beta, \bm \beta'}[\sum_{j=1}^p (1-\sigma_j^2)(\beta_j^2+\beta_j'^{2}+2\sigma_j\beta_j\beta'_j) + \sum_{k=p+1}^{p'}\beta_k'^2].
\end{equation}
When $p=p'$, Cauchy-Schwarz inequality guarantees that $2\beta_j\beta'_j \le \beta_j^2+\beta_j'^2$, where the equation holds when $ \beta_j= \beta'_j$. It follows that
\begin{equation}
\label{equ:representative-conv}
\begin{aligned}
\td(Z, Z'; \gS^*) \xrightarrow{a.s.} & \sup_{\bm \beta, \bm \beta'} \sum_{j=1}^p(1-\sigma_j)(1+\sigma_j)^2(\beta_j^2+\beta_j'^2) \\
& =\max_{j=1,\cdots,p} 2(1-\sigma_j)(1+\sigma_j)^2.
\end{aligned}
\end{equation}

When $p<p'$, if there exists an $i$ such that $(1-\sigma_j)(1+\sigma_j)^2\ge 1$, then the optimal $\bm \beta$ satisfies $\beta'_{p+1}=\cdots=\beta'_{p'}=0$, so we can achieve the same result as (\ref{equ:representative-conv}). Thus, we prove the results in Theorem \ref{thm:convergence}. \qed

\subsection{Proof of Corollary \ref{thm:td-restricted}}

Here, we use the same notation as the previous section. Based on the results in Theorem \ref{thm:convergence}, we now consider restricted task sets. In $\gS_r = \{Y = \bm \alpha A^{-\frac{1}{2}}Z + \bm \alpha' C^{-\frac{1}{2}} Z' : \left\| \bm \alpha \right\|_2 \leq 1, \left\| \bm \alpha' \right\|_2 \leq 1, \bm \alpha \mathbf u_j = 0 \text{ and } \bm \alpha' \mathbf v_j = 0 \text{ if } j > r\}$, we have $\beta_j = \beta'_j=0$ for $j > r$. It yields

\begin{equation}
\begin{aligned}
\td(Z, Z'; \gS_r) \xrightarrow{a.s.} & \sup_{\bm \beta, \bm \beta'} \sum_{j=1}^r(1-\sigma_j)(1+\sigma_j)^2(\beta_j^2+\beta_j'^2) \\
& =\max_{j=1,\cdots,r} 2(1-\sigma_j)(1+\sigma_j)^2.
\end{aligned}
\end{equation}

which concludes the proof. \qed

\subsection{The Invairance Properties of the TD metric}

For a general loss function $\ell(h_W(\mathbf z), y)$ that is Lipschitz, strongly convex in $h_W(\mathbf z)$ and satisfies $\ell(y, y)=0$, we will show that it also satisfy the two invariance properties mentioned in Section 4. When training the output head's parameter $\hat W$ from

\begin{equation}
    \hat W=\argmin_{W}\frac{1}{n} \sum_{i=1}^{n}\ell(h_W(\mathbf{z}_i), y_i)),
\end{equation}

there exists only one optimal $h_W(\mathbf{z}_i)$ due to the convexity of the loss function $\ell$. Denote the output set of $h_{W}(\mathbf{Z})$ as $S_{\mathbf{z}}:=\{h_{\hat W}(\mathbf z)| \forall W\}$. It is easy to see that, if there is a linear transformation directly applied to $\mathbf z$ by $h_W(\mathbf z)$, the output set will be invariant to isotropic scaling and orthogonal transformation. $S_{\mathbf z}$ = $S_{\beta \mathbf z}$ for all $\beta$ and $S_{\mathbf z}=S_{Q\mathbf z}$ for all unitary matrix $Q$. Although the optimization algorithm may find different $W$, the output $h_W(\mathbf z_i)$ remains unchanged for each data. Therefore, the TD metric defined as $\td(Z,Z';Y)=\frac{1}{n} \sum_{i=1}^n d(h_{W}(\mathbf z_i), h_{W'}(\mathbf z'_i))$ is invariant to isotropic scaling and orthogonal transformation.

\subsection{Empirical Analysis of the Distribution of Singular Values}

\begin{figure}[htb]
  \centering 
  \includegraphics[width=5cm]{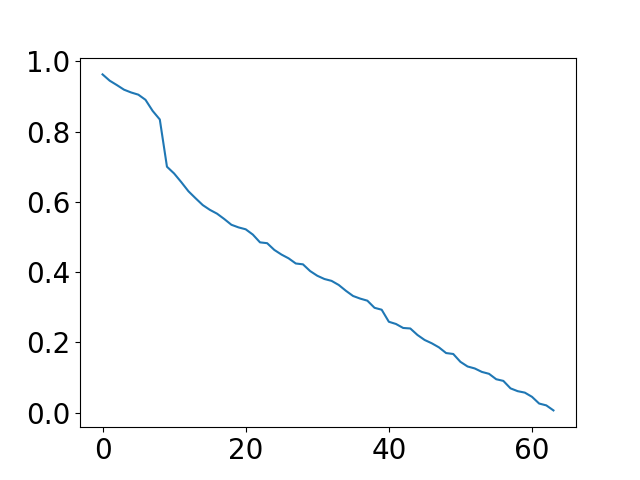}\\
  \caption{Singular values of $D$}
  \label{fig:singular}
\end{figure}

Previous theoretical analysis shows that in the linear probing setting, the TD metric is closely related to the distribution of singular values of $D=A^{-\frac{1}{2}}BC^{-\frac{1}{2}}$. If all the singular values are large, then the two representations will have similar performance on all the tasks in the representative set. If there exists a small singular value, then there exists a direction that is close to one representation matrix's row space but nearly orthogonal to the other representation matrix's row space. On a task $y\in \gS^*$ in this direction, these two representations will have drastically different performance. When all the singular values are small, these two representations will perform differently on all tasks in the representative task set. Now, we empirically look at the distribution of singular values of $D$ in practice.

We train two ResNet32 models on the Cifar-10 training set with different random initializations (i.e. seeds), extract the representation matrix on the Cifar-10 test set, and calculate the empirically estimator $\hat{ D}=(ZZ^{\top})^{-\frac{1}{2}}ZZ'^{\top}(Z'Z'^{\top})^{-\frac{1}{2}}$. The training hyperparameters are the same as in C.1. We plot the distribution of the singular values of $\hat D$ in Figure \ref{fig:singular}. It can be seen that the singular values larger than 0.6 take only a small fraction among all the singular values. Most singular values are small, and there exist singular values close to 0. Thus, the two representations will have similar performance on the task related to the large singular value and have drastically different performance on the task related to the small singular value. Therefore, a universal similarity index may not reveal all the difference, and one may consider evaluating the two representations on a set of downstream tasks of interest. When evaluated on the restricted task set we proposed in Section 4.2, the TD is 0.288 on $\gS_1$ and 0.649 on $\gS_5$. When using TD in practice, the task set can also be selected as various tasks we want to deal with using the pre-trained feature extractor. 

In Section \ref{sec:cca-d}, Theorem \ref{thm:mean_convergence} provides that the CCA index obtains a universal index by taking the average over all tasks in the representative set or over the squares of the singular values. Since most of the singular values are small and the performance is different on most tasks, the CCA value shall be large and varies little for different models. In Section D's experimental results, the CCA value is always larger than 0.6 and remains stable when trained with different techniques of data augmentation and different training strategies.

\section{Transferred Discrepancy versus the Maximum Match, CCA and CKA}

\label{sec:tdvs}
\subsection{Main Results for the Maximum Match and CKA}\label{append:compare}

In Section \ref{sec:cca-d} we discussed the relationship between CCA and TD. Here we present our main results for the maximum match and CKA, which show that both of them are closely related to the matrix $D$.

\paragraph{Maximum Match} 

Max-match measures the intersection between two subspaces of the row spaces of $Z$ and $Z'$. Denote the row vectors of $A^{-\frac{1}{2}}Z$ by $Z_{row}=\{z^{(1)},\cdots,z^{(p)}\}$ and the row vectors of $C^{-\frac{1}{2}}Z'$ by $Z'_{row}=\{z'^{(1)},\cdots,z'^{(p')}\}$. Let $\hat{Z}_{row}$ and $\hat{Z'}_{row}$ be the largest subspaces of $Z_{row}$ and $Z'_{row}$ such that they are $\epsilon$-close to each other ($\epsilon \ge 0$). Here closeness means that for any $z^{(i)} \in \hat{Z}_{row}$, $\min_{z' \in \mspan(\hat{Z'}_{row})}\left\| z^{(i)}-z' \right\|_2 \leq \sqrt{n}\epsilon$ and vice versa. The max-match similarity index is defined as

\begin{equation}
   S_{max-match}(\epsilon)= \frac{|\hat{Z}_{row}|+|\hat{Z'}_{row}|}{p+p'} 
\end{equation}

Compared with the original definition in \cite{NIPS2018_8167}, we normalize $Z$ and $Z'$ by $A^{-\frac{1}{2}}$ and $C^{-\frac{1}{2}}$, so that they are at the same scale with respect to $\epsilon$. We also add $\sqrt{n}$ in the definition of $\epsilon$-closeness so that the result is meaningful as $n \rightarrow \infty$. The following theorem provides the relationship between $S_{max-match}$ and $D$:

\begin{thm} 
\label{thm:maxmatch} 
Let $s = \E_{Z,Z'}[S_{max-match}(\epsilon)]$. Denote the singular values of $D$ by $\sigma_1 \geq\cdots\geq \sigma_p$. Let $k$ be the largest integer such that

\begin{equation}
    \sigma_1^2+\cdots+\sigma_k^2 \geq k(1-\epsilon^2),
\end{equation}

Then we have

\begin{equation}
\label{equ:smm}
    s \le \frac{2k}{p+p'}.
\end{equation}
\end{thm}

In \cite{NIPS2018_8167}, the authors found that the maximum match similarity is very small when $\epsilon \leq 0.3$. Figure \ref{fig:singular} shows that large singular values account for a very small fraction of all singular values of $D$. Thus, it is a natural consequence of Theorem \ref{thm:maxmatch} that $S_{max-match}(\epsilon)$ should be very small. This metric investigate the two feature matrix under linear transformation, and \cite{Liang2020Knowledge} extended it to non-linear transformation using several convolution and activation layers.

\paragraph{Centered Kernel Alignment (CKA)}  \cite{pmlr-v97-kornblith19a} proposed CKA based on dot product and its extension in reproducing Hilbert spaces. Particularly, the linear CKA similarity index is defined as
\begin{equation}
S_{CKA} (Z^{\top}Z,Z'^{\top}Z')=\frac{\|Z'Z^{\top}\|_F^2}{\|ZZ^{\top}\|_F\|Z'Z'^{\top}\|_F}.
\end{equation}

\begin{thm}
The expectation of CKA is
\begin{equation}
    \E [\cka(Z^{\top}Z, Z'^{\top}Z^{\top})]=\frac{\tr(DCD^{\top}A)}{\sqrt{\tr(A^2)}\sqrt{\tr(C^2)}}.
\end{equation}
\end{thm}

This result is straightforward from the definition. It indicates that the CKA gets a universal similarity index by re-weighting the matrix $D$ with matrices $A$ and $C$. 

\subsection{Lemma for Theorem \ref{thm:mean_convergence}}

\begin{lem}
\label{lem:mean_conv}

If $\mathbf{x} \in \R^{p}$ is uniformly distributed in the unit ball, then $\E {\mathbf{x}_1^2} = \frac{1}{p+2}$.

\end{lem}

\emph{Proof} \quad We directly compute this expectation:

\begin{equation}
\begin{aligned}
\E \mathbf{x}_1^2 = \frac{\int_{-1}^{1}\int_{\mathbf{x}_2^2+\cdots+\mathbf{x}_p^2\le 1-\mathbf{x}_1^2} \mathbf{x}_1^2 dV d\mathbf{x}_1}{\int_{-1}^{1}\int_{\mathbf{x}_2^2+\cdots+\mathbf{x}_p^2\le 1-\mathbf{x}_1^2} dV d\mathbf{x}_1}.
\end{aligned}
\end{equation}

Denote the volume of the unit ball in $\R^{p}$ by $V_p$. We have $\int_{\mathbf{x}_2^2+\cdots+\mathbf{x}_p^2\le t^2} dV = t^{p-1}V_{p-1}$. Denote the Beta function by $B(P,Q)$. For the numerator, let $t^2 = 1-\mathbf{x}_1^2$, and we have

\begin{equation}
\begin{aligned}
\int_{-1}^{1}\int_{\mathbf{x}_2^2+\cdots+\mathbf{x}_p^2\le 1-\mathbf{x}_1^2} \mathbf{x}_1^2 dV d\mathbf{x}_1 &=  \int_{-1}^{1} t^{p-1}V_{p-1} \mathbf{x}_1^2 dV d\mathbf{x}_1 \\
&= V_{p-1}\int_{-1}^{1} (1-\mathbf{x}_1^2)^{\frac{p-1}{2}} \mathbf{x}_1^2 d\mathbf{x}_1 \\
& = V_{p-1} \left( \int_{-1}^{1} (1-\mathbf{x}_1^2)^{\frac{p-1}{2}} d\mathbf{x}_1 - \int_{-1}^{1} (1-\mathbf{x}_1^2)^{\frac{p+1}{2}} d\mathbf{x}_1 \right) \\
& = V_{p-1} \left( \int_{-\frac{\pi}{2}}^{\frac{\pi}{2}} \cos^p \theta d\theta - \int_{-\frac{\pi}{2}}^{\frac{\pi}{2}} \cos^{p+2} \theta d\theta \right)\\
&= V_{p-1} \left( B(\frac{p+1}{2}, \frac{1}{2})-B(\frac{p+3}{2}, \frac{1}{2})\right)\\
&=\frac{1}{p+2}V_{p-1}B(\frac{p+1}{2}, \frac{1}{2}).
\end{aligned}
\end{equation}

Similarly, the denominator is equal to $V_{p-1}B(\frac{p+1}{2}, \frac{1}{2})$. Thus,
$\E\mathbf{x}_1^2 = \frac{1}{p+2}$. \qed

\subsection{Proof of Theorem \ref{thm:mean_convergence}}

Suppose $\hat D$ is defined as the empirical estimator of $D$, i.e. $\hat D=(Z'Z'^{\top})^{-\frac{1}{2}}Z'Z^{\top}(ZZ^{\top})^{-\frac{1}{2}}$. Denote the singular values of $\hat D$ as $1\ge \hat \sigma_1 \ge ... \ge \hat \sigma_p \ge 0$. Similar to Eqn (\ref{equ:td-1task}) and Eqn (\ref{equ:suppthm1}),
we have

\begin{equation}
    \begin{aligned}
    & \E_{\bm \alpha,\bm \alpha'}\left[\td(Z, Z'; Y=\bm \alpha \hat A^{-\frac{1}{2}}Z+\bm \alpha' \hat C^{-\frac{1}{2}}Z')\right]  \\
    = &\E_{\bm \alpha, \bm \alpha'} \left[\bm \alpha(I_p - \hat D \hat D^{\top})\bm \alpha^{\top} + \alpha'(I_{p'} - \hat D^{\top}\hat D)\alpha'^{\top} + 2\alpha(\hat D-\hat D\hat D^{\top}\hat D)\alpha'^{\top}\right] \\
    = & \E_{\bm \beta, \bm \beta'}\left[\sum_{j=1}^p (1-\hat \sigma_j^2)(\beta_i^2+\beta_i'^2+2\hat \sigma_i\beta_i\beta'_i) + \sum_{k=p+1}^{p'}\beta_k'^2\right].
\end{aligned}
\end{equation}

Here we also change the variable $\bm \alpha, \bm \alpha'$ to $\bm \beta, \bm \beta'$ with $\bm \beta = \bm \alpha \hat U$ and $\bm \beta' = \bm \alpha' \hat V$. $\hat U$ and $\hat V$ are the orthogonal matrices in $\hat D$'s singular value decomposition $\hat D = \hat U \hat \Sigma \hat V^{\top}$. Since $\bm \alpha$ is uniformly distributed in the unit ball in $\R^p$, so does $\bm \beta$. Similarly, $\bm \beta'$ is uniformly distributed in the unit ball in $\R^{p'}$. Thus, $\E \beta_j \beta'_j=0$. It follows by Lemma \ref{lem:mean_conv} that $\E \beta_j^2=\frac{1}{p+2}$ for all $j \in [p]$, and $\E \beta_k'^2=\frac{1}{p'+2}$ for all $k \in [p']$. Therefore,

\begin{equation}
\E_{\bm \alpha,\bm \alpha'}[\td(Z, Z'; Y=\bm \alpha \hat A^{-\frac{1}{2}}Z+\bm \alpha' \hat C^{-\frac{1}{2}}Z')] = \frac{p-\sum_{j=1}^p \hat \sigma_j^2}{p+2} + \frac{p'-\sum_{j=1}^p \hat \sigma_j^2}{p'+2}.
\end{equation}

Similar to Eqn (\ref{equ:cca_sing}), $R_{CCA}^2=\frac{\|\hat D\|_F^2}{p}=\frac{\sum_{j=1}^p \hat \sigma_j^2}{p}$. Thus, we have

\begin{equation}
\E_{\alpha, \alpha'}[\td(Z,Z'; Y=\bm \alpha \hat A^{-\frac{1}{2}}Z+\bm \alpha' \hat C^{-\frac{1}{2}}Z')]=\frac{2p}{p+2}(1-R_{CCA}^2).
\end{equation}

This results shows that the calculation of $R_{CCA}$ is equivalent to averaging on the TD over all linearly realizable tasks. This averaging produces a downstream-task-agnostic metric but dismissing many information. Consequently, the CCA varies little in all the experiments and is insensitive to the change of training strategies. \qed

\subsection{Lemma for Theorem \ref{thm:maxmatch}}

\begin{lem}
\label{lma:thm3}

Suppose that $A\in \R^{n\times n}$ is a symmetric semi-positive definite matrix. Let the eigenvalues of $A$ be $0\le \lambda_1 \le \lambda_2 \le \cdots \le \lambda_n$. For a constant $\epsilon \geq 0$, let $k$ be the largest integer such that $\lambda_1+\cdots+\lambda_k \leq k \epsilon$. Suppose that $\mathbf{v}_1,\cdots,\mathbf{v}_m \in \R^n$ are unit vectors orthogonal to each other, i.e. $\|\mathbf v_i\|_2=1$ for all $i$ and $\mathbf v_i \mathbf v_j^{\top}=0$ for all $i \neq j$. If $\mathbf v_iA \mathbf v_i^{\top}\le \epsilon$ for all $i=1,\cdots,m$, then $m \leq k$.
\end{lem}

\emph{Proof} \quad There exists a unitary matrix $Q$ such that $A=Q\Lambda Q^{\top}$, where $\Lambda = \diag\{\lambda_1, \lambda_2, \cdots, \lambda_n\}$. Let $\mathbf{u}_i = Q \mathbf{v}_i$, $i=1,\cdots,m$. Due to the property of unitary matrices, $\mathbf{u}_i$ are unit vectors orthogonal to each other. Denote $\mathbf{u}_i=(\mathbf u_i^1, \mathbf u_i^2, \cdots, \mathbf u_i^n)$. Then, we have

\begin{equation}
\epsilon \geq \mathbf v_i A \mathbf v_i^{\top} = \mathbf u_i \Lambda \mathbf u_i^{\top}= \sum_{j=1}^n \lambda_j (\mathbf u_i^j)^2 .
\end{equation}

Let $x_j=\sum_{i=1}^m (\mathbf{u}_i^j)^2$. Since $\mathbf{u}_i$ are unit vectors orthogonal to each other, we have $0 \leq x_j \leq 1$ and $x_1+\cdots+x_n = m$. Thus,

\begin{equation}
m \epsilon \geq \sum_{i=1}^m \sum_{j=1}^n \lambda_j (\mathbf{u}_i^j)^2 = \sum_{j=1}^n \lambda_j x_j \geq \lambda_1+\cdots+\lambda_m.
\end{equation}

Therefore, by the definition of $k$ we have $m \leq k$. \qed

\subsection{Proof of Theorem \ref{thm:maxmatch}}

Let $\hat{Z}_{row}$ and $\hat{Z'}_{row}$ be the two largest subspaces described in the theorem. For each $z^{(i)} \in \hat{Z}_{row}$, there exists an $\mathbf{a}_i\in \R^{p'}$ such that $\left\| z^{(i)}-\mathbf{a}_iC^{-\frac{1}{2}}Z'\right\|_2 \leq \sqrt{n}\epsilon$. Let $1_i=(0,\cdots,0,1,0,\cdots,0)$ be a $p$-dimensional row vector whose $i^{th}$ element is 1 and the rest are 0. Thus, $\mathbf{b}_i=(1_i, -\mathbf{a}_i)$ satisfy

\begin{equation}
    \mathbf{b}_i\begin{bmatrix}A^{-\frac{1}{2}}ZZ^{\top}A^{-\frac{1}{2}}&A^{-\frac{1}{2}}ZZ'^{\top}C^{-\frac{1}{2}}\\C^{-\frac{1}{2}}Z'Z^{\top}A^{-\frac{1}{2}}&C^{-\frac{1}{2}}Z'Z'^{\top}C^{-\frac{1}{2}}\end{bmatrix}\mathbf{b}_i^{\top} = \|z^{(i)}-\mathbf{a}_iC^{-\frac{1}{2}}Z'\|_2^2 \leq n\epsilon^2.
\end{equation}

By taking the expectation we have

\begin{equation}
    \epsilon^2 \geq \mathbf{b}_i \begin{bmatrix} \mI_p & D \\ D^{\top} & \mI_{p'} \end{bmatrix} \mathbf{b}_i^{\top} = 1_i (I_p - DD^{\top}) 1_i^{\top} + \left\| \mathbf{a}_i-1_i D \right\|_2^2
\end{equation}

Thus, for every $z^{(i)} \in \hat{Z}_{row}$ we have

\begin{equation}
    1_i (I_p - DD^{\top}) 1_i^{\top} \leq \epsilon^2
\end{equation}

The eigenvalues of $I_p - DD^{\top}$ are $0 \leq 1-\sigma_1^2 \leq \cdots \leq 1-\sigma_p^2$. Since $1_i$ are unit vectors orthogonal to each other, using Lemma \ref{lma:thm3} we have $|\hat{Z}_{row}| \leq k$. Similarly, for each $z'^{(j)} \in \hat{Z'}_{row}$,

\begin{equation}
1_j(I_{p'}-D^{\top}D)1_j^{\top}\le\epsilon^2.
\end{equation}

Consequently, $|\hat{Z'}_{row}| \leq k$. Thus, $s \leq \frac{2k}{p+p'}$. \qed

\section{Experimental Settings}
\label{sec:exp-detail}

All experimental settings are listed in this section and results are reported in Section D.

\subsection{Settings of Sections 5.1 and 5.2}

\begin{figure}[htb]
    \centering
    \includegraphics[width=1\linewidth]{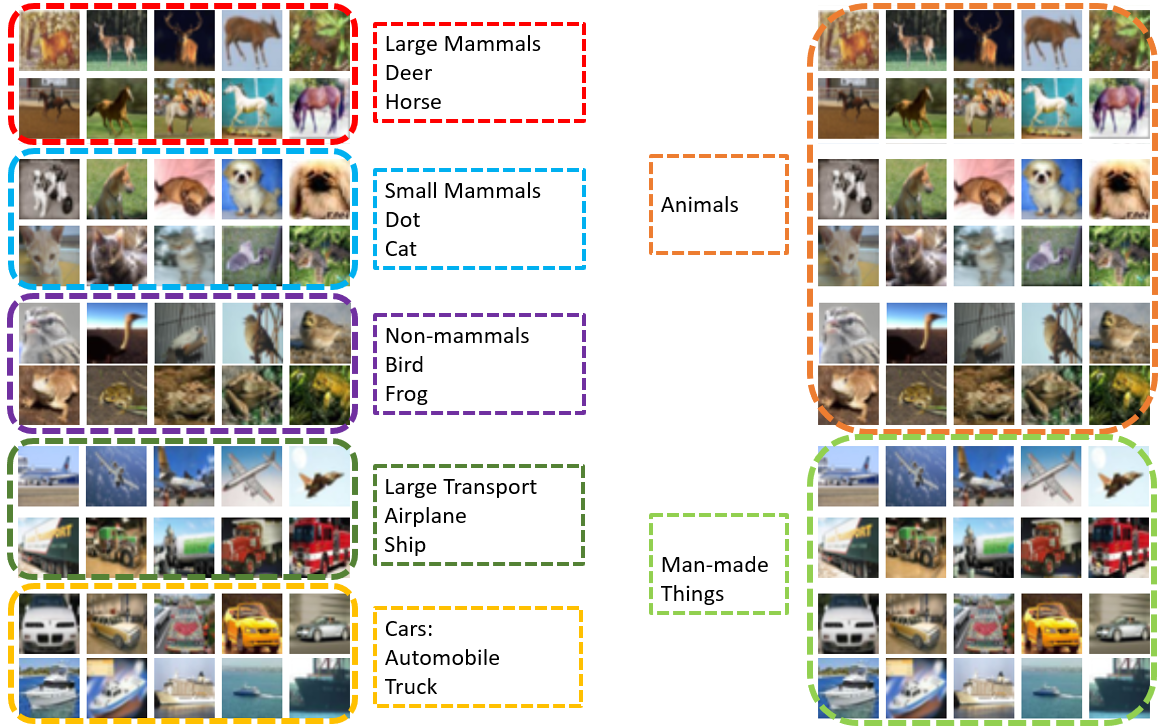}
    \caption{A visualization of Cifar-2 (left) and Cifar-5 (right).}
    \label{fig:cifar2/5}
\end{figure}

\paragraph{The Design of Upstream Tasks and Downstream Tasks}
We design several tasks based on the CIFAR-10 dataset and the SVHN dataset. For the CIFAR-10 dataset, we manually group the labels to design three different tasks: Cifar-10, Cifar-5 and Cifar-2, each of which can be used either as an upstream task or as a downstream task. The Cifar-10 task uses the original CIFAR-10 labels, and Cifar-5 and Cifar-2 are built by semantically regrouping those 10 classes into 5 and 2 categories, as shown in Table \ref{tab:cifar10labels}. Figure \ref{fig:cifar2/5} provides a visualization of the categories of Cifar-5 and Cifar-10. For the SVHN dataset, we directly use its original labels. 

\begin{table}[htb]
\centering
\caption{Two tasks induced from CIFAR-10: Cifar-2 task(first table) and Cifar-5 task(second table).}
\label{tab:cifar10labels}
\begin{tabular}{|l|l|}
\hline
Cifar-2 Category & Classes \\ \hline
Man-made transport & Airplane, Automobile, ship, truck \\ \hline
animals & Bird, cat, deer, dog, horse, frog \\ \hline
\end{tabular}
\vspace{.05in}
\begin{tabular}{c}
       \\
\end{tabular}
\begin{tabular}{|l|l|}
\hline
Cifar-5 Category & Classes \\ \hline
Cars & Automobile, truck \\ \hline
Large mammals & Deer, horse \\ \hline
Medium mammals & Cat, dog \\ \hline
Large transport & Ship, airplane \\ \hline
Non-mammals & Frog, bird \\ \hline
\end{tabular}
\end{table}

\paragraph{Architecture} We use ResNet-32 \cite{he2016deep} for all tasks. 

\paragraph{Upstream Task Training (Feature Extractor)} For any upstream task, we train the model for 200 epochs with SGD and the batch size is set to 128. The learning rate is initially set to 0.1 and decayed by 0.1 at epochs 60 and 120. The momentum is 0.9, and the weight decay rate is 5e-4. Unless explicitly stated, random cropping and random horizontal flipping are applied for data augmentation. After training, we remove the network's last fully-connected layer (output head) and take the remaining network as the feature extractor. 

\paragraph{Downstream Task Training} For any downstream task, we add a linear layer with softmax on top of the feature extractor as the output head. We freeze the feature extractor and only fine-tune the output head using the training set of the downstream task. The output head is trained for 50 epochs with SGD and the batch size is set to 128. All other hyperparameters are set to the same value as the upstream task training. The model's performance on the test set is reported.

\paragraph{More TD Metrics for Evaluation}
Recall that the TD metric is defined as $d(h_{W_1}(\mathbf z),h_{\hat W_2}(\mathbf z))$ in (\ref{equ:diff_metric}), where $d(\cdot,\cdot)$ is a distance metric. We consider two TD metrics defined with different $d(\cdot,\cdot)$:

\begin{itemize}
    \item Soft Distance: $\td_{soft}(Z, Z')=\gW_1 (h_{W_1}(Z), h_{W_2}(Z'))=\frac{1}{n}\sum_{i=1}^n\frac{1}{2}\|h_{W_1}(Z_i)-h_{W_2}(Z'_i)\|_1,$
    \item Hard Distance: $\td_{hard}(Z, Z')=\frac{1}{n}\sum_{i=1}^n \mathbf{1}_{\argmax h_{W_1}(Z_i) = \argmax h_{W_2}(Z'_i)}$.
\end{itemize}

$\td_{hard}$ is used in the paper as $\td_{cls}$, which directly measures the fraction of samples on which the two models have different predictions. $\td_{soft}$ defined with the $l_1$ distance is equivalent to the Wasserstein distance between the two probability densities. Both $\td_{hard}$ and $ \td_{soft}$ are metrics between [0,1], and small numbers indicate more similar representations. Although $\td_{soft}$ and $\td_{hard}$ are based on different distance metrics, they exhibit similar trends in most experiments. 

\paragraph{Tasks Based on CIFAR-100}

We also conduct experiments with the CIFAR-100 dataset, in which each sample has two levels of labels: a fine label (100 classes) and a coarse label (20 superclasses) as listed in Table \ref{tab:cifar100labels1}. The two levels of original labels produce two tasks: C 100 and C 20. We further design two tasks, C 10 and C 4, by semantically regrouping the samples based on the labels' relationship extracted from the WordTree in ImageNet \cite{krizhevsky2012imagenet}. Details are shown in Table \ref{tab:cifar100labels2}. 

\begin{table}[htb]
\caption{Two original labelings from CIFAR-100: 100 classes and 20 superclasses.}
\label{tab:cifar100labels1}
    \centering
    \begin{tabular}{|l|l|}
    \hline
        Superclass (C 20 Category) & Classes \\ \hline
        aquatic mammals & beaver, dolphin, otter, seal, whale \\ \hline
        fish & aquarium fish, flatfish, ray, shark, trout \\ \hline
        flowers & orchids, poppies, roses, sunflowers, tulips \\ \hline
        food containers & bottles, bowls, cans, cups, plates \\ \hline
        fruit and vegetables & apples, mushrooms, oranges, pears, sweet peppers \\ \hline
        household electrical devices & clock, computer keyboard, lamp, telephone, television \\ \hline
        household furniture & bed, chair, couch, table, wardrobe \\ \hline
        insects & bee, beetle, butterfly, caterpillar, cockroach \\ \hline
        large carnivores & bear, leopard, lion, tiger, wolf \\ \hline
        large man-made outdoor things & bridge, castle, house, road, skyscraper \\ \hline
        large natural outdoor scenes & cloud, forest, mountain, plain, sea \\ \hline
        large omnivores and herbivores & camel, cattle, chimpanzee, elephant, kangaroo \\ \hline
        medium-sized mammals & fox, porcupine, possum, raccoon, skunk \\ \hline
        non-insect invertebrates & crab, lobster, snail, spider, worm \\ \hline
        people & baby, boy, girl, man, woman \\ \hline
        reptiles & crocodile, dinosaur, lizard, snake, turtle \\ \hline
        small mammals & hamster, mouse, rabbit, shrew, squirrel \\ \hline
        trees & maple, oak, palm, pine, willow \\ \hline
        vehicles 1 & bicycle, bus, motorcycle, pickup truck, train \\ \hline
        vehicles 2 & lawn-mower, rocket, streetcar, tank, tractor \\ \hline
    \end{tabular}
    
\end{table}

\begin{table}[htb]

    \centering
    \caption{CIFAR-100-induced task: C 4 (first table) with four classes and C 10 (second table) with 10 classes.}
    \label{tab:cifar100labels2}
    \begin{tabular}{|l|l|}
    \hline
        C 4 Category & Superclass \\ \hline
        mammals & Aquatic mammals, large carnivores,\\
        & large omnivores and herbivores,\\
        & small mammals, medium-sized mammals, people \\ \hline
        Non-mammals & Fish, Reptiles, insects, non-insect invertebrates,  \\ \hline
        Man-made things & Vehicles 1, vehicles 2, Food containers,\\ & household electrical devices, \\
        & household furniture, Large man-made outdoor things \\ \hline
        Natural things and plants & Trees, flowers, fruit and vegetables,\\& Large natural outdoor scenes \\ \hline
    \end{tabular}
    
    \begin{tabular}{cc}
          & \\
    \end{tabular}
    
    \vspace{0.05in}
    \begin{tabular}{|l|l|}
    \hline
        C 10 Category & Superclass \\ \hline
        Aquatic animals & Aquatic mammals, fish \\ \hline
        Large animals & large carnivores, large omnivores and herbivores \\ \hline
        Medium and small mammals & small mammals, medium-sized mammals  \\ \hline
        Vehicles & Vehicles 1, vehicles 2. \\ \hline
        Other animals & Reptiles, insects, non-insect invertebrates \\ \hline
        People & people \\ \hline
        Plants & Trees, flowers, fruit and vegetables \\ \hline
        Household & Food containers, household electrical devices, \\& household furniture \\ \hline
        Large man-made outdoor things & Large man-made outdoor things \\ \hline
        Large natural outdoor scenes & Large natural outdoor scenes \\ \hline
    \end{tabular}
    
\end{table}

\subsection{Settings of Section 5.3}
\label{exp:different_seed}

\paragraph{Data Augmentation} We study three data augmentation techniques: random flipping, random cropping (padding = 4), and adding Gaussian noise ($\sigma=0.1$).

\paragraph{Learning Rate Schedule} We train the feature extractor for 100 epochs with three learning rate schedules. In the ``Small LR + without decay" schedule, the learning rate is fixed at 0.01; in the ``Large LR + without decay" schedule, the learning rate is fixed at 0.1; and in the ``Large LR + with decay" schedule, the learning rate is 0.1 in the first 50 epochs and 0.01 in the last 50 epochs. 

\paragraph{Adversarial training} Adversarial training \cite{madry2017towards} is one of the successful training method to improve the robustness of neural networks to adversarial attacks. The main idea to add adversarial examples into the training set during training to improve robustness. We further investigat how adversarial training affect TD robustness.
We use PGD-7 to generate adversarial examples and set the perterbation to $\frac{8}{255}$ and step size to $\frac{2}{255}$ during training. When trained on CIFAR-10, this setting of the adversarial training can achieve 79.22\% clean accuracy and 48.23 \% robustness under the same level of attacks.

Apart from the three factors above, we also study how other factors affect TD robustness.

\paragraph{Batch Size} We train the feature extractor with different batch sizes. We set the batch size to 32, 64, 128, 256, and 512, following the common practice of setting the batch size to a power of 2.

\paragraph{Architectures} We investigate how the model's width and depth affect the TD robustness. For the ResNet models\cite{he2016deep}, we study models with different depths including ResNet20, ResNet32, ResNet44, ResNet56 and ResNet110, and models with different widths (Wide ResNet) such as 2xResNet32, 5xResNet32, and 10xResNet32. We also experiments on VGG models\cite{simonyan2014very} including VGG 13-bn, VGG 16-bn, and VGG 19-bn.

\paragraph{Upstream Tasks} 

We study how the choice of upstream task affects the representation learned by the feature extractor. We use the four tasks based on the CIFAR-100 dataset: C 4/10/20/100.

\section{Experimental Results}
\subsection{A Sanity Check}


The sanity check results are reported in Section 5.1 in Table \ref{supp:sanity}, including $\td_{soft}$ for each row. It is clear that the first three rows have smaller $D_{CCA}$, $D_{CKA}$ and $\td_{hard}$ values than the other rows, which indicates that the features learned from Cifar-2/5/10 are more similar, but are very different from the features learned from SVHN. 

\begin{table}[htb]
    \centering
     \caption{The sanity check results}
    \begin{tabular}{|l|l|l|l|l|l|l|}
    \hline
        Model 1 & Model 2 & Downstream & $D_{CCA}$ & $D_{CKA}$ & $\td_{soft}$ & $\td_{hard}$ \\ \hline
        Cifar-10 & Cifar-5 & \multirow{6}{*}{Cifar-10} & \cellcolor[RGB]{251, 158, 147}0.7642 & \cellcolor[RGB]{253, 217, 212}0.3024 & \cellcolor[RGB]{253, 213, 208}0.3344 & \cellcolor[RGB]{254, 228, 225}0.2158 \\ \cline{1-2}\cline{ 4-7}
        Cifar-10 & Cifar-2 &  & \cellcolor[RGB]{251, 149, 138}0.8323 & \cellcolor[RGB]{252, 187, 180}0.5330 & \cellcolor[RGB]{252, 170, 160}0.6717 & \cellcolor[RGB]{253, 192, 185}0.4958\\ \cline{1-2}\cline{4-7}
        Cifar-5 & Cifar-2 &  & \cellcolor[RGB]{251, 153, 142}0.8030 & \cellcolor[RGB]{253, 197, 191}0.4530 & \cellcolor[RGB]{252, 191, 184}0.5057 & \cellcolor[RGB]{253, 192, 185}0.4944  \\ \cline{1-2}\cline{ 4-7}
        Cifar-10 & SVHN &  & \cellcolor[RGB]{250, 138, 125}0.9218 & \cellcolor[RGB]{250, 132, 118}0.9713 & \cellcolor[RGB]{251, 147,135}0.8528 & \cellcolor[RGB]{251, 153, 142}0.8001  \\ \cline{1-2}\cline{ 4-7}
        Cifar-5 & SVHN &  & \cellcolor[RGB]{250, 131, 117}0.9793 & \cellcolor[RGB]{250, 132, 118}0.9710 & \cellcolor[RGB]{251, 159, 149}0.7521 & \cellcolor[RGB]{251, 153, 142}0.8049 \\ \cline{1-2}\cline{ 4-7}
        Cifar-2 & SVHN &  & \cellcolor[RGB]{250, 139, 126}0.9163 & \cellcolor[RGB]{250, 131, 117}0.9782 & \cellcolor[RGB]{253, 192, 185}0.5000 & \cellcolor[RGB]{251, 151, 140}0.8163 \\ \hline
    \end{tabular}
    \label{supp:sanity}
    \vspace{.05in}
   
\vspace{-.15in}
\end{table}

\subsection{Does Initialization Affect Learned Features?}

In this section, we further study whether models trained from different random initializations learn similar representations using a variety of downstream tasks. The results are reported in Table \ref{tab:supp-downstream}.

As shown by the results, $\td_{soft}$ and $\td_{hard}$ exhibit similar trends and output different values for different downstream tasks. When we train the model on Cifar-5, both models provide consistent predictions on Cifar-2, but the variance gets much bigger on Cifar-10. Furthermore, it can be observed that from C 4 to C 100, as the task becomes more difficult, the difference between the two representations increases. On C 4, the two models trained on Cifar-5 disagree on 32.77\% of the test samples. When evaluated on C 100, however, the same two models disagree on as much as 80.05\% of the data. The variation of TD as demonstrated in the table implies that a reasonable metric measuring the difference between two representations should take the downstream tasks into consideration.

\begin{table}[htb]
    \centering
    \caption{Similarity of the models learned with different initializations.}
    \label{tab:supp-downstream}
    \begin{tabular}{lllllll}
    \shline
    \specialrule{0em}{1.2pt}{1.2pt}
        Model 1 & Model 2  & Downstream & $D_{CCA}$ & $D_{CKA}$ & $\td_{soft}$ & $\td_{hard}$
        \\ 
        \specialrule{0em}{1.2pt}{1.2pt} \hline \specialrule{0em}{1.2pt}{1.2pt}
        \multirow{8}{*}{Cifar-5} & \multirow{8}{*}{Cifar-5} & Cifar-10 & \multirow{3}{*}{0.6961} & \multirow{3}{*}{0.0835} & 0.1617 & 0.2139  \\ 
         & & Cifar-5 & & & 0.0200 & 0.0442  \\ 
         & & Cifar-2 & & & 0.0124 & 0.0109 \\ 
         \specialrule{0em}{1.2pt}{1.2pt} \cline{3-7} \specialrule{0em}{1.2pt}{1.2pt}
         & & C 100 & \multirow{4}{*}{0.8003} & \multirow{4}{*}{0.4210} & 0.3113 & 0.8005 \\
         & & C 20 & & & 0.2599 & 0.6377 \\
         & & C 10 & & & 0.2347 & 0.5010 \\
         & & C 4 & & & 0.1845 & 0.3277 \\
         \specialrule{0em}{1.2pt}{1.2pt} \cline{3-7} \specialrule{0em}{1.2pt}{1.2pt}
         &  & SVHN & 0.8663 & 0.6940 & 0.1765 & 0.5312 \\
         \specialrule{0em}{1.2pt}{1.2pt} \hline
         \specialrule{0em}{1.2pt}{1.2pt}
        \multirow{8}{*}{Cifar-2} & \multirow{8}{*}{Cifar-2} & Cifar-10 & \multirow{3}{*}{0.6931} &  \multirow{3}{*}{0.0402} & 0.1489 & 0.3745 \\ 
         & & Cifar-5 & & & 0.1281 & 0.2631\\ 
         & & Cifar-2 & & & 0.0168 & 0.0164  \\
          \specialrule{0em}{1.2pt}{1.2pt} \cline{3-7} \specialrule{0em}{1.2pt}{1.2pt}
         & & C 100 & \multirow{4}{*}{0.7533} & \multirow{4}{*}{0.3036} & 0.1762 & 0.7651\\
         & & C 20 & & & 0.1595 & 0.6018 \\
         & & C 10 & & & 0.1485 & 0.4259 \\
         & & C 4 & & & 0.1243 & 0.2644 \\
         \specialrule{0em}{1.2pt}{1.2pt} \cline{3-7} \specialrule{0em}{1.2pt}{1.2pt}
        &  & SVHN & 0.8330 & 0.6305 & 0.1296 & 0.4157 \\
         \specialrule{0em}{1.2pt}{1.2pt} \shline 
    \end{tabular}
\end{table}

Now we try to answer whether the features learned by two models trained from different initializations are similar. We want to emphasize that the question depends on how we evaluate the features, i.e., which downstream task set we choose. Among all the tasks, Cifar-2/5/10 are the most correlated. The models trained on Cifar-5 can achieve 80\% accuracy on Cifar-10, indicating that the features can be successfully transferred. However, even on such a correlated task, the two models disagree on 21.39\% of the test samples. Thus, the features captured by models trained from different initializations are not the same. The TD metric allows a practitioner to choose the downstream tasks of their interest, and both $\td_{hard}$ and $\td_{soft}$ have clear practical meanings: $\td_{hard}$ shows the fraction of samples that two models disagree on while $\td_{soft}$ measures the difference between predictions in terms of the likelihood indicated by the softmax output.

\subsection{TD Robustness and Its Application to Training Strategy Evaluation}

\paragraph{Data Augmentation}
We calculate TD robustness under various configurations of data augmentation's techniques, including random flipping (F), random cropping (C), and Gaussian additive noise (G). Full results are reported in Table \ref{tab:supp-dataaug}.

\begin{table}[htb]
    \centering
    \caption{The effect of data augmentation: We study random flipping (F), random cropping (C) and adding Gaussian noise (G). +/- indicates whether the data augmentation method is applied.}
    \label{tab:supp-dataaug}
    \begin{tabular}{lllllllll}
    \shline
    \specialrule{0em}{1.2pt}{1.2pt}
        Upstream & Downstream & F & C & G & $D_{CCA}$ & $D_{CKA}$ & $\td_{soft}$ & $\td_{hard}$ \\
        \specialrule{0em}{1.2pt}{1.2pt}
        \hline
        \specialrule{0em}{1.2pt}{1.2pt}
        \multirow{8}{*}{Cifar-5} & \multirow{8}{*}{Cifar-10} & - & - & - & 0.7677 & 0.0866 & 0.2199 & 0.2812 \\
        && + & - & - & 0.7552 & 0.0887 & 0.1692 & 0.2739 \\
        && - & + & - & 0.7306 & {0.0758} & 0.1614 & 0.2260 \\
        && + & + & - & {0.7285} & 0.0833 & {0.1593} & {0.2150} \\
        && - & - & + & 0.7783 & 0.1310 & 0.1940 & 0.3107 \\
        && + & - & + & 0.7668 & 0.1319 & 0.1900 & 0.2790 \\
        && - & + & + & 0.7603 & 0.1370 & 0.2051 & 0.2773 \\
        && + & + & + & 0.7500 & 0.1319 & 0.1976 & 0.2542 \\
        \specialrule{0em}{1.2pt}{1.2pt}
        \shline
    \end{tabular}
    \vspace{.05in}
    
    \vspace{-.15in}
    \label{tab:data-aug-full}
\end{table}

In the results, $\td_{soft}$ and $\td_{hard}$ share similar trends. The results show that both random flipping and random cropping have positive effects on TD robustness while the Gaussian additive noise decrease TD robustness. For $\td_{hard}$, random cropping has a more significant effect than random flipping. Applying random cropping can reduce the value of $\td_{hard}$ from 0.2812 to 0.2260, while flipping only reduces it to 0.2739. For $\td_{soft}$, random flipping and random cropping have comparable effects. On the other hand, both $\td_{hard}$ and $\td_{soft}$ increase when Gaussian noise is added. 
The results reveal that TD robustness matches previous understandings that flipping and cropping help learn a good representation while Gaussian noise may harm the generalization \cite{shorten2019survey, chen2020simple}.

\paragraph{Learning Rate Schedule}

Full results are reported in Table \ref{tab:lr-schedule}. All four metrics show that the models trained with the ``Large LR + with decay" schedule capture more similar features than the models trained with the other two schedules. These observations coincide with the fact that initialing with a large learning rate and then decaying it in the middle of training helps improve the quality of the representation. A theoretical analysis of this phenomenon can be found in \cite{NIPS2019_9341}. 

\begin{table}[htb]
    \centering
    \caption{The effect of learning rate schedule.}
    \label{tab:lr-schedule}
    \begin{tabular}{lllllll}
    \shline
    \specialrule{0em}{1.2pt}{1.2pt}
    Upstream & Downstream & Schedule & $D_{CCA}$ & $D_{CKA}$ & $\td_{soft}$ & $\td_{hard}$ \\
    \specialrule{0em}{1.2pt}{1.2pt}
        \hline
        \specialrule{0em}{1.2pt}{1.2pt}
        \multirow{3}{*}{Cifar-5} & \multirow{3}{*}{Cifar-10} & Small LR + without decay & 0.7739 & 0.1502 & 0.2118 & 0.2521 \\
        & & Large LR + without decay & 0.7417 & 0.2855 & 0.2470 & 0.2770 \\
        & & Large LR + with decay & 0.7264 & 0.1232 & 0.1896 & 0.2313 \\
        \specialrule{0em}{1.2pt}{1.2pt}
        \shline
    \end{tabular}
\end{table}

\paragraph{Adversarial training} In Table \ref{tab:supp-adv}, we report the results for standard training and adversarial training. When the training strategy changes from standard training to adversarial training, $\td_{hard}$ decreases from 0.2139 to 0.1823, and $\td_{soft}$ drops from 0.1617 to 0.0883. It can be concluded that although adversarial training lowers the classification accuracy of the model, it improves TD robustness and makes models trained from different initializations learn more similar features. This conclusion coincides with the widespread belief that adversarial training helps models capture features that align better with human perception \cite{tsipras2019robustness, ilyas2019adversarial, engstrom2019learning}.

\begin{table}[htb]
    \centering
    \caption{The difference between training with standard methods and adversarial training.}
    \label{tab:supp-adv}
    \begin{tabular}{lllllll}
    \shline
    \specialrule{0em}{1.2pt}{1.2pt}
        Upstream & Downstream & Training & $D_{CCA}$ & $D_{CKA}$ & $\td_{soft}$ & $\td_{hard}$ \\ \specialrule{0em}{1.2pt}{1.2pt} \hline \specialrule{0em}{1.2pt}{1.2pt}
        \multirow{2}{*}{Cifar-5} & \multirow{2}{*}{Cifar-10}& Standard  & 0.6961 & 0.0835 & 0.1617 & 0.2139 \\ 
        \specialrule{0em}{1.2pt}{1.2pt}
         & & Adversarial & 0.6339 & 0.0717   & {0.0883} & {0.1823} \\
         \specialrule{0em}{1.2pt}{1.2pt} \shline
    \end{tabular}
\end{table}

\paragraph{Batch Size} The results on how batch size affects TD robustness are reported in Table \ref{tab:supp-batch}. All four metrics verify that batch size 128 is the optimal setting for the default learning rate schedule. In practice, batch size 128 achieves the highest accuracy, and batch size 256 decreases the average accuracy by 1\%.

\begin{table}[htb]
    \centering
    \caption{The effect of batch size.}
    \label{tab:supp-batch}
    \begin{tabular}{lllllll}
    \shline
    \specialrule{0em}{1.2pt}{1.2pt}
    Upstream & Downstream & Batch Size & $D_{CCA}$ & $D_{CKA}$ & $\td_{soft}$ & $\td_{hard}$ \\
    \specialrule{0em}{1.2pt}{1.2pt}
        \hline
        \specialrule{0em}{1.2pt}{1.2pt}
        \multirow{5}{*}{Cifar-5} & \multirow{5}{*}{Cifar-10} & 32 & 0.7422 & 0.1484 & 0.2201 & 0.2858 \\
        & & 64 & 0.7244 & 0.1271 & 0.1772 & 0.2436 \\
        & & 128 & {0.6961} & 0.0835 & {0.1617} & {0.2139} \\
        & & 256 & 0.7054 & {0.0690}& 0.1704 & 0.2142 \\
        & & 512 & 0.7252 & 0.0744 & 0.1646 & 0.2214 \\
        \specialrule{0em}{1.2pt}{1.2pt}
        \shline
    \end{tabular}
\end{table}

\paragraph{Number of Training Epochs}

\begin{figure}[htb]
    \centering
    \includegraphics[width=0.8\linewidth]{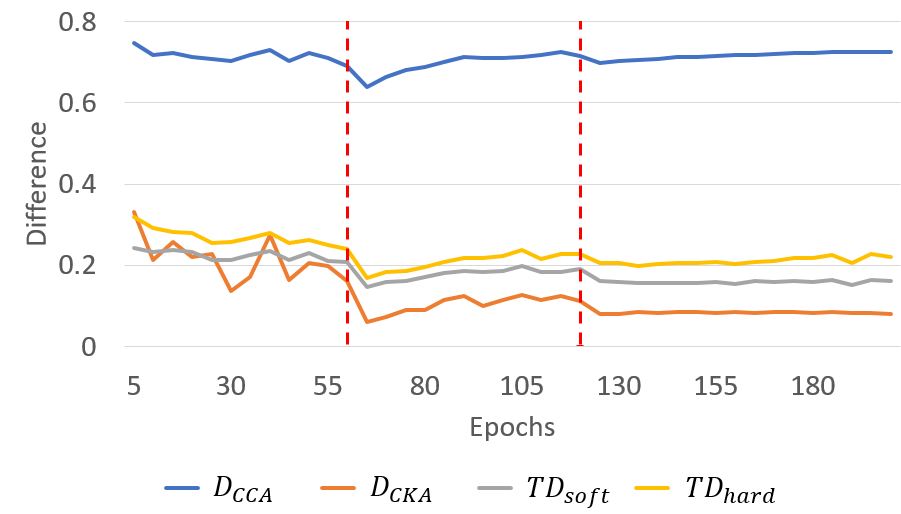}
    \caption{The changes in difference metrics during training.}
  \label{fig:epochs}
\end{figure}

We plot TD robustness during training in Figure \ref{fig:epochs}. The two red vertical lines indicate the epochs when the learning rate is decayed with factor 0.1. The plot discloses the properties of training that the performance curve does not manifest. For instance, after learning rate decay, all curves of four metrics drop significantly and then rise up again, indicating that the models are overfitting the training samples at a lower learning rate.

\paragraph{Architectures}

We also experiment with different architectures, including ResNet and VGG of different depths and widths, and report the results in Table \ref{tab:archi}. For ResNet models, when the depth increases from 20 to 110, $\td_{hard}$ also decreases steadily, and the disagreement decreases from 22.53\% to 19.76\%. $\td_{soft}$ also shares similar trends, but the minimum point occurs for ResNet56 models. $D_{CCA}$ first goes up and then down from ResNet32 to ResNet 44 and to ResNet56. $D_{CKA}$ indicates that ResNet56 models capture similar features than other models. In practice, it is also believed that increasing the depth is helpful for the quality of the representation and the model's generalization, aligning with the trends of $\td_{hard}$. In our experiments, when transferred to Cifar-10, the ResNet110 trained on Cifar-5 can reach 83\% average accuracy while the number for ResNet20 is only 76\%. The increased depth helps learn better features for transferring to Cifar-10.

As we use ResNet of different widths, both $\td_{hard}$ and $\td_{soft}$ decreases when the width increases. Simply doubling the width from 1xResNet32 to 2xResNet32 results in a significant reduction in $\td_{hard}$. Intuitively, when the width increases, the representation's dimension also increases, and more features are learned. In this case, two representations are more likely to have similar features. Although it is widely believed that the depth has a more significant impact on the performance than the width, our experiments show that the width may be more important for the similarity between two representations than the depth.

\begin{table}[htb]
    \centering
    \caption{The effect of model architectures.}
    \begin{tabular}{lllllll}
    \shline
    \specialrule{0em}{1.2pt}{1.2pt}
    Upstream & Downstream & Architecture & $D_{CCA}$ & $D_{CKA}$ & $\td_{soft}$ & $\td_{hard}$ \\
    \specialrule{0em}{1.2pt}{1.2pt}
    \hline
    \specialrule{0em}{1.2pt}{1.2pt}
    \multirow{12}{*}{Cifar-5} & \multirow{12}{*}{Cifar-10} & ResNet20 & 0.7414 & 0.0982 & 0.1617 & 0.2253 \\ 
   &  &  ResNet32 & 0.6961  & 0.0835 & 0.1617 & 0.2139 \\ 
    & &   ResNet44 & 0.7023 & 0.0736  & 0.1583 & 0.2021 \\
    &  &   ResNet56 & 0.6997 & 0.0710 & {0.1561} & 0.2015 \\
    & &   ResNet110 & 0.6887 & 0.0736 & 0.1627 & {0.1976} \\ 
    \specialrule{0em}{1.2pt}{1.2pt}
    \cline{3-7}
    \specialrule{0em}{1.2pt}{1.2pt}
    & & 1xResNet32 & 0.6961 & 0.0835 & 0.1617 & 0.2139 \\
    & & 2xResNet32 & 0.6917 & 0.0473 & 0.1315 & 0.1580\\
    & & 5xResNet32 & 0.7038 & 0.0349 & 0.1082 & 0.1530 \\
    & & 10xResNet32 & 0.7125 & 0.0422 & 0.1285 & 0.1478 \\
    \specialrule{0em}{1.2pt}{1.2pt}
    \cline{3-7}
    \specialrule{0em}{1.2pt}{1.2pt}
    & &   VGG13 & 0.7411 & 0.0415 & 0.1297 & {0.3753} \\ 
    & &   VGG16 & 0.7767 & 0.0419 & 0.1408 & 0.4567 \\
    & &   VGG19 & 0.7974 & 0.0455 & {0.1179} & {0.3838} \\
    \specialrule{0em}{1.2pt}{1.2pt}
    \shline
    \end{tabular}
    \label{tab:archi}
\end{table}

\begin{table}[htb]
    \centering
        \caption{The effect of different Upstream tasks}
    \begin{tabular}{llllll}
    \shline
    \specialrule{0em}{1.2pt}{1.2pt}
        Upstream & Downstream & $D_{CCA}$ & $D_{CKA}$ & $\td_{soft}$ & $\td_{hard}$ 
        \\
        \specialrule{0em}{1.2pt}{1.2pt}
        
        \hline
        \specialrule{0em}{1.2pt}{1.2pt}
        C 100 & \multirow{4}{*}{Cifar-10} & 0.7379& 0.4352 & 0.3028 & {0.3340} \\
        C 20 & & 0.8221 & 0.5287 & 0.3307 & 0.4362 \\ 
        C 10 & & 0.8153 & 0.4750& 0.3242 & 0.4386 \\ 
        C 4 & & 0.8016& 0.4002  & 0.3072 & 0.4953 \\ 
        \specialrule{0em}{1.2pt}{1.2pt}
        \hline
        \specialrule{0em}{1.2pt}{1.2pt}
        C 100 & \multirow{4}{*}{Cifar-5} & 0.7379& 0.4352 & 0.2124 & {0.2169} \\
        C 20 & & 0.8221 & 0.5287 & 0.2511 & 0.2807 \\ 
        C 10 & & 0.8153 & 0.4750 & 0.2472 & 0.2673 \\ 
        C 4 & & 0.8016  & 0.4002 & 0.2541 & 0.3179\\ 
        \specialrule{0em}{1.2pt}{1.2pt}
        \shline
    \end{tabular}
    \label{tab:upstream}
\end{table}

\paragraph{Upstream Tasks}

Using which upstream task to train the feature extractor is the core question in representation learning. In our experiment, we study four upstream tasks: C 4/10/20/100. We evaluate the results on Cifar-10 and Cifar-5 and report the results in Table \ref{tab:upstream}. From C 4 to C 100, the upstream task becomes more difficult and contains more information. In practice, the models trained on harder upstream tasks have higher accuracy on both the Cifar-5 and Cifar-10 downstream tasks. $\td_{soft}$, $\td_{hard}$, and $D_{CCA}$ all show that the models trained on the C 100 task learn the most similar features in the table while $D_{
CKA}$ outputs the smallest number for models trained with C 4.

\end{document}

%% file: math_commands.tex
\usepackage{amsmath,amsfonts,bm}

\def\eqref#1{equation~\ref{#1}}

\def\1{\bm{1}}

\def\mI{{\bm{I}}}

\DeclareMathAlphabet{\mathsfit}{\encodingdefault}{\sfdefault}{m}{sl}
\SetMathAlphabet{\mathsfit}{bold}{\encodingdefault}{\sfdefault}{bx}{n}

\def\gS{{\mathcal{S}}}

\def\gW{{\mathcal{W}}}
\def\gX{{\mathcal{X}}}

\newcommand{\pdata}{p_{\rm{data}}}

\newcommand{\E}{\mathbb{E}}

\newcommand{\R}{\mathbb{R}}

\DeclareMathOperator*{\argmax}{arg\,max}
\DeclareMathOperator*{\argmin}{arg\,min}